\documentclass{article} %
\usepackage{iclr2024_conference_arxiv,times}
\iclrfinalcopy

\usepackage{amsmath,amsfonts,bm}

\def\eqref#1{equation~\ref{#1}}

\def\1{\bm{1}}

\DeclareMathAlphabet{\mathsfit}{\encodingdefault}{\sfdefault}{m}{sl}
\SetMathAlphabet{\mathsfit}{bold}{\encodingdefault}{\sfdefault}{bx}{n}

\usepackage{hyperref}

\usepackage{url}            %
\usepackage{booktabs}       %
\usepackage{amsfonts}       %
\usepackage{nicefrac}       %
\usepackage{microtype}      %
\usepackage{xcolor}         %

\usepackage{colortbl}
\usepackage{tablefootnote}
\usepackage{graphicx}
\usepackage{multirow}
\usepackage{enumitem}
\usepackage{wrapfig}
\usepackage{subcaption}

\DeclareUnicodeCharacter{00D7}{-}
\DeclareUnicodeCharacter{2212}{-}
\DeclareUnicodeCharacter{00B0}{-}

\newcommand*\samethanks[1][\value{footnote}]{\footnotemark[#1]}

\title{Take a Step Back: Evoking Reasoning via Abstraction in Large Language Models}
\title{Step-Back Prompting Enables Reasoning Via Abstraction in Large Language Models}
\title{Take a Step Back: Evoking Reasoning via Abstraction in Large Language Models}

\author{
Huaixiu Steven Zheng\thanks{Equal Contribution} \quad
Swaroop Mishra\samethanks{} \quad
Xinyun Chen \quad
Heng-Tze Cheng \\
\textbf{
Ed H. Chi \quad
Quoc V Le \quad
Denny Zhou
}
\\
[1ex]
 Google DeepMind \\
}

\renewcommand{\headrulewidth}{1pt}
\fancypagestyle{firstpage}{
  \lhead{\begin{picture}(0,0)\put(0,-3){\includegraphics[width=0.25\linewidth]{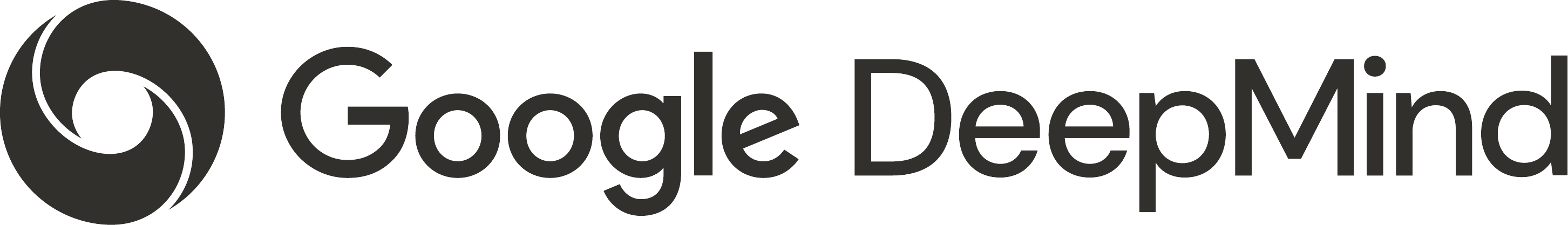}}\end{picture}}
}

\newcommand{\name}{\textsc{Step-Back Prompting}}

\iclrfinalcopy %
\begin{document}

\maketitle
\thispagestyle{firstpage}
\vspace{-1em}
\begin{abstract}
We present \name{}, a simple prompting technique that enables LLMs to do abstractions to derive high-level concepts and first principles from instances containing specific details. Using the concepts and principles to guide reasoning, LLMs significantly improve their abilities in following a correct reasoning path towards the solution. 
We conduct experiments of \name{} 
with PaLM-2L, GPT-4 and Llama2-70B models, and observe substantial performance gains on various challenging reasoning-intensive tasks including STEM, Knowledge QA, and Multi-Hop Reasoning. For instance, \name{} improves PaLM-2L performance on MMLU (Physics and Chemistry) by $7\%$ and $11\%$ respectively, TimeQA by $27\%$, and MuSiQue by $7\%$.
\end{abstract}

\textit{The purpose of abstraction is not to be vague, but to create a new semantic level in which one can be absolutely precise.  — Edsger W. Dijkstra}\\
\section{Introduction}

The field of natural language processing (NLP) is witnessing a ground-breaking revolution because of the Transformer-based ~\citep{vaswani2017attention} large language models (LLMs) ~\citep{devlin2018bert, raffel2020exploring, brown2020language, anil2023palm}. Scaling up the model size and pre-training corpus~\citep{hoffmann2022training, chowdhery2022palm} has brought remarkable improvement in model capabilities and sample efficiency with insights from the scaling law~\citep{kaplan2020scaling, hoffmann2022training}, as well as emergent abilities~\citep{wei2022emergent} such as multi-step reasoning~\citep{wei2022chain, zhou2022least} and instruction following~\citep{mishra2022cross, wei2021finetuned}. 

\begin{figure*}[!h]
\centerline{\includegraphics[width=\linewidth]{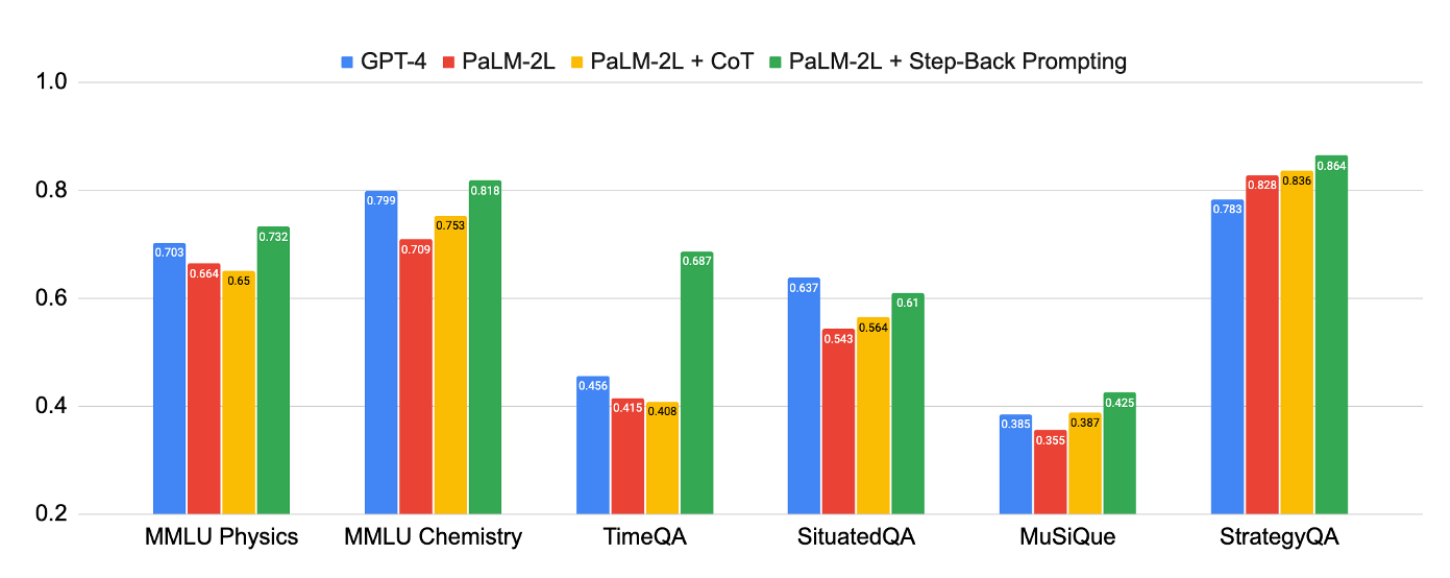}}
\caption{Strong Performance of \name{}: our proposed Abstraction-and-Reasoning scheme leads to a substantial improvement in a wide range of challenging tasks in STEM, Knowledge QA and Multi-Hop Reasoning requiring complex (often multi-hop) reasoning.
}
\label{fig:result_summary}
\end{figure*}

Despite the great advancements, complex multi-step reasoning remains challenging for even the state-of-the-art LLMs. \cite{lightman2023let} show that process-supervision with step-by-step verification is a promising remedy to improve the correctness of intermediate reasoning steps. Techniques such as Chain-of-Thought \citep{wei2022chain} were introduced to produce a coherent series of intermediate reasoning steps to increase the success rate of following the right decoding path. Inspired by the fact that when faced with challenging tasks humans often step back and do abstractions to arrive at high-level principles to guide the process, we propose \name{} to ground reasoning on abstractions to reduce the chance of making errors in the intermediate reasoning steps.

\begin{figure*}[!h]
\centerline{\includegraphics[width=\linewidth]{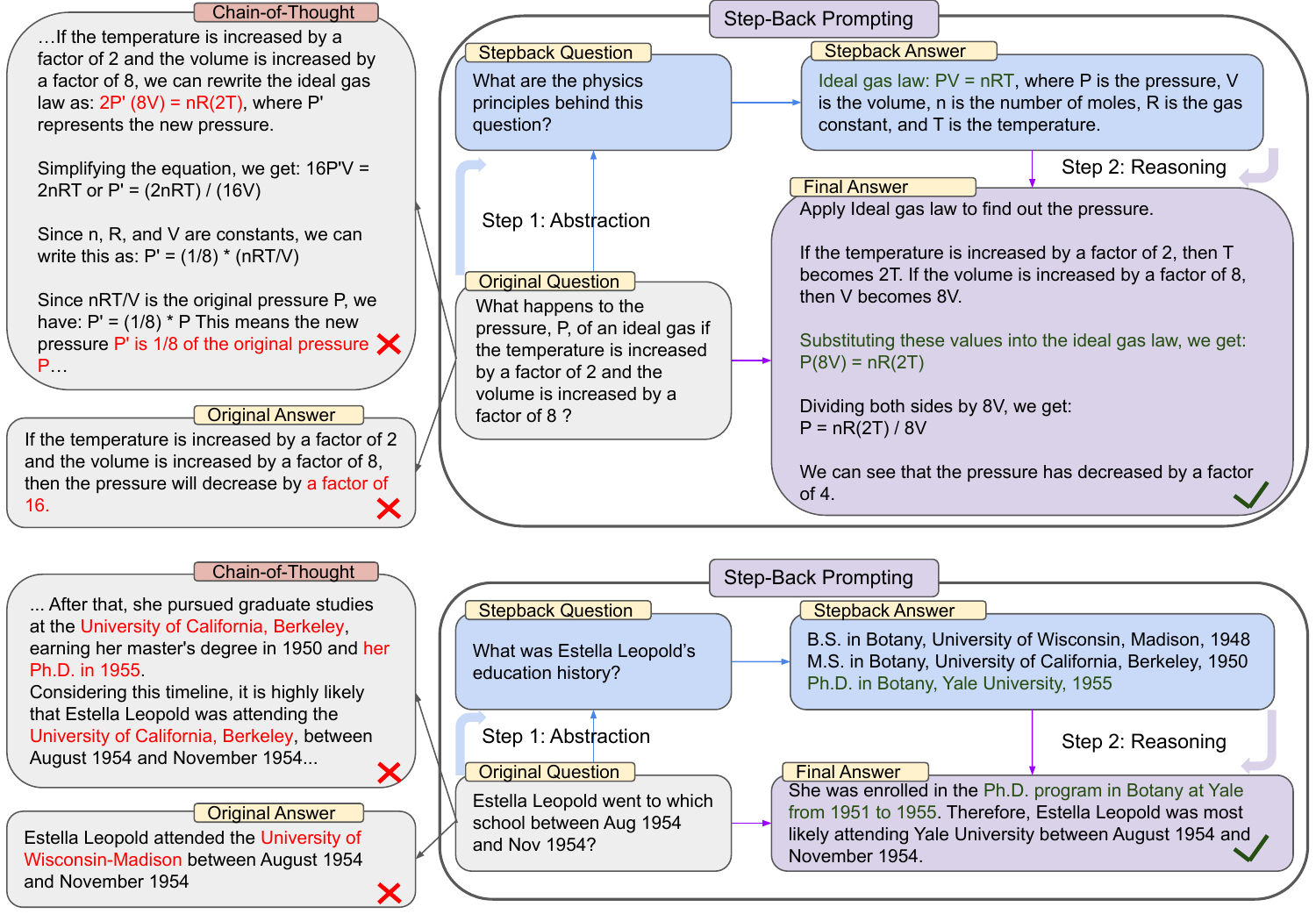}}
\caption{Illustration of \name{} with two steps of Abstraction and Reasoning guided by concepts and principles.
\textit{Top}: an example of MMLU high-school physics~\citep{hendrycks2020measuring} where the first principle of Ideal Gas Law is retrieved via abstraction.
\textit{Bottom}: an example from TimeQA~\citep{chen2021dataset} where the high-level concept of education history is a result of the abstraction. 
\textit{Left}: PaLM-2L~\citep{anil2023palm} fails to answer the original question. Chain-of-Thought prompting \citep{wei2022chain, kojima2022large} ran into errors during intermediate reasoning steps (highlighted as red). \textit{Right}: PaLM-2L~\citep{anil2023palm} successfully answers the question via \name{}.
}
\label{fig:sbt_diagram}
\end{figure*}

Among many of the cognitive skills, abstraction \citep{lachmy2022} is ubiquitous to humans' ability to process vast amounts of information and derive general principles. For example, Kepler compressed thousands of measurements into Kepler’s three laws of planetary motion, which precisely describe the orbits of planets around the Sun \citep{russell1964kepler}. In critical decision-making, humans find abstraction to be helpful since it provides a broader view of the environment. This work explores how LLMs can tackle complex tasks involving many low-level details through a two-step process of abstraction-and-reasoning. The first step is to show LLMs how to step back through in-context learning -- prompting them to derive high-level abstractions such as concepts and principles for a specific example. The second step is to leverage the reasoning ability to reason on top of the high-level concepts and principles. We use few-shot exemplar demonstrations to execute \name{} on LLMs.

We experiment across a range of tasks involving domain specific reasoning such as Physics and Chemistry, knowledge-intensive question answering requiring factual knowledge, multi-hop commonsense reasoning. We observe significant performance improvements (up to $27\%$) in PaLM-2L~\citep{anil2023palm} demonstrating the efficacy of \name{} in tackling complex tasks, which are otherwise challenging due to the amount of details needed for reasoning. Figure~\ref{fig:result_summary} shows a summary of all the key results presented in this paper. Some the tasks are very challenging: both PaLM-2L and GPT-4 achieve only $\sim40\%$ accuracy on TimeQA and MuSiQue. Chain-of-Thought prompting leads to a minor improvement on a few tasks, while \name{} improves the performance of PaLM-2L across the board: $7\%$ and $11\%$ on MMLU Physics and Chemistry, $27\%$ on TimeQA, and $7\%$ on MuSiQue.

We conduct a variety of analyses and find that \name{} leads to strong performance improvements (up to $36\%$) over chain-of-thought (CoT) prompting~\citep{wei2022chain} and ``take-a-deep-breath'' (TDB) prompting~\citep{yang2023large}. We perform a qualitative evaluation where we find that Step-Back fixes a large portion of errors of the base model (up to $\sim$ $40\%$) while introducing a small portion of new errors (max $\sim$ $12\%$). We also conduct an error analysis and find that majority of the errors made by \name{} is attributed to the intrinsic limitations of reasoning capabilities of LLMs while abstraction skills are relatively easy to demonstrate to LLMs, pointing out the direction for future improvements of methods alike \name{}.

\section{\name}
\name{} is motivated by the observation that many tasks contain a lot of details, and it is hard for LLMs to retrieve relevant facts to tackle the task. As shown in the first example (top) in Figure~\ref{fig:sbt_diagram}, for a Physics question of ``\textit{What happens to the pressure, P, of an ideal gas if the temperature is increased by a factor of 2 and the volume is increased by a factor of 8 ?}'', the LLM can deviate from the first principle of Ideal Gas Law when reasoning directly on the question.
Similarly, a question of ``\textit{Estella Leopold went to which school between Aug 1954 and Nov 1954?}'' is very hard to address directly given the detailed time range constraint. In both cases, asking a step-back question helps the model to solve the problem effectively. 

We define a step-back question as a derived question from the original question at a higher level of abstraction. For instance, instead of directly asking ``\textit{which school Estella Leopold went to during a specific period}'', a step-back question (Figure~\ref{fig:sbt_diagram} bottom) would ask about the ``\textit{education history}'', which is a high-level concept encompasses the original question. Answering the step-back question of ``\textit{Estella Leopold's education history}'' in this case will provide all the necessary information to reason about ``\textit{which school Estella Leopold went to during a specific period}''. The premise is that the step-back question is typically much easier. Grounding the reasoning on top of such abstractions helps to avoid reasoning errors in the intermediate steps such as the example shown in Figure~\ref{fig:sbt_diagram} (left) from Chain-of-Thought.
In short, \name{} consists two simple steps:
\begin{itemize}[leftmargin=*, nolistsep]
\setlength{\itemsep}{1mm}
    \item \textbf{Abstraction}: Instead of addressing the question directly, we first prompt the LLM to ask a generic step-back question about a higher-level concept or principle, and retrieve relevant facts about the high-level concept or principle. The step-back question is unique for each task in order to retrieve the most relevant facts.
    \item \textbf{Reasoning}: Grounded on the facts regarding the high-level concept or principle, the LLM can reason about the solution to the original question. We term this as \textit{Abstraction-grounded Reasoning}.
\end{itemize}

In the following sections, we present an empirical study of \name{} on a range of challenging tasks covering STEM, Knowledge QA, and Multi-Hop Reasoning involving complex reasoning.

\section{Experimental Setup}
Here we define the tasks and models we experiment with. We also describe our evaluation metric and the baselines we consider.

\subsection{Tasks} We experiment with the following diverse tasks: (a) STEM, (b) Knowledge QA, and (c) Multi-Hop Reasoning. We describe below the datasets we consider (see Appendix~\ref{app:dataset_details} for more details).

\begin{itemize}
    \item \textbf{STEM}: We evaluate MMLU and GSM8K for STEM tasks. MMLU \citep{hendrycks2020measuring} contains a series of benchmarks across diverse domains to evaluate the model's language understanding. We consider the high school physics and chemistry portions of MMLU because of the deep reasoning involved.
    \item \textbf{Knowledge QA}: We consider TimeQA~\citep{chen2021dataset} since it contains complex queries that require challenging time-sensitive knowledge. We also experiment with SituatedQA~\citep{zhang2021situatedqa}, another challenging open-retrieval QA dataset requiring the model to answer questions given temporal or geographical contexts.
    \item \textbf{Multi-Hop Reasoning}: We experiment with MuSiQue~\citep{trivedi2022musique}, a hard multihop reasoning dataset created via composable pairs of single-hop questions, and StrategyQA \citep{geva2021did} with open-domain questions that demand some strategy to solve.
\end{itemize}

\subsection{Models}
We use the following state-of-the-art LLMs: instruction-tuned PaLM-2L~\citep{anil2023palm}, GPT-4~\citep{gpt4}, and Llama2-70B \citep{touvron2023llama}. 

\subsection{Evaluation}
Conventional evaluation metrics such as accuracy, F1 score have limitations specifically for evaluating the generations of state-of-the-art LLMs since these models often generate long-form answers which are hard to capture. We instead conduct an evaluation using the PaLM-2L model where we few-shot prompt the model to identify equivalence between target answers and the model predictions. Few-shot examples, prompts and other details used for this evaluation are in Appendix~\ref{app:eval}.
\subsection{Baseline Methods}
\begin{itemize}
    \item \textbf{PaLM-2L, PaLM-2L 1-shot}: PaLM-2L is either queried directly with the question or has a single demonstration exemplar of question-answer included in the prompt.
    \item \textbf{PaLM-2L + CoT, PaLM-2L + CoT 1-shot}: PaLM-2L model is queried with zero-shot CoT prompting \citep{kojima2022large}: ``\textit{Let's think step by step}'' is appended to the question. For 1-shot, One demonstration example of a question and answer pair is provided in the prompt, where the answer is in the style of CoT \citep{wei2022chain}.
    \item \textbf{PaLM-2L + TDB}: Zero-shot prompting with ``\textit{Take a deep breath and work on this problem step-by-step.}'' \citep{yang2023large} prepended to the question.
    \item \textbf{PaLM-2L + RAG}: For Sections~\ref{sec:timeqa} and \ref{sec:multihop_reasoning}, we use retrieval-augmented generation (RAG) where the retrieved passage is used as context by the LLM. 
    \item \textbf{GPT-4 and Llama2-70B}: we run GPT-4 and Llama2-70B on MMLU tasks for all methods. In addition, we also run GPT-4 on all baselines for all tasks.
\end{itemize}
We do not use RAG for STEM tasks, because of the inherent reasoning nature of the tasks contrary to the other fact-seeking datasets. All inferences are done using greedy decoding.

\begin{table}%
\vspace{-2mm}
\caption{Strong performance of \name{} on MMLU tasks across three model families. CoT: zero-shot Chain of Thought prompting \citep{kojima2022large}, TDB: Take a Deep Breath prompting \citep{yang2023large}.}
\label{table:mmlu_physics}
\begin{center}
\begin{tabular}{l|c|c}
\toprule
Method & MMLU Physics & MMLU Chemistry \\
\midrule 
PaLM-2L & 66.4\% (0.8\%) & 70.9\% (0.9\%)\\
PaLM-2L 1-shot & 64\% (1.6\%) & 75.6\% (0.4\%)\\
PaLM-2L + CoT & 65\% (2\%) & 75.3\% (1.5\%)\\
PaLM-2L + CoT 1-shot & 61.5\% (1.8\%) & 76.6\% (1\%)\\ 
PaLM-2L + TDB & 65.7\% (0.7\%) & 73.8\% (1.1\%)\\
PaLM-2L + Step-Back (ours) & \textbf{73.2\%} (1.9\%) & \textbf{81.8}\% (1.4\%)\\
\midrule 
GPT-4 & 69.4\% (2.0\%) & 80.9\% (0.7\%)\\
GPT-4 1-shot & 78.4\% (2.4\%) & 80.5\% (1.6\%)\\
GPT-4 + CoT & 82.9\% (0.5\%) & 85.3\% (1.0\%)\\
GPT-4 + CoT 1-shot & 79.3\% (1.0\%) & 82.8\% (0.5\%)\\ 
GPT-4 + TDB & 74.4\% (4.0\%) & 81.5\% (1.3\%)\\
GPT-4 + Step-Back (ours) & \textbf{84.5\%} (1.2\%) & \textbf{85.6}\% (1.4\%)\\
\midrule 
Llama2-70B & 51.9\% (3.6\%) & 55.7\% (2.1\%)\\
Llama2-70B 1-shot & 57.3\% (1.6\%) & 58.5\% (2.5\%)\\
Llama2-70B + CoT & 59.3\% (2.0\%) & 64.1\% (1.2\%)\\
Llama2-70B + CoT 1-shot & 59.6\% (2.0\%) & \textbf{68.1\%} (1.4\%)\\ 
Llama2-70B + TDB & 60.4\% (2.1\%) & 63.6\% (1.9\%)\\
Llama2-70B + Step-Back (ours) & \textbf{64.8\%} (1.5\%) & 66.7\% (1.6\%)\\
\bottomrule
\end{tabular}
\end{center}
\end{table}

\section{STEM}
We evaluate \name{} on STEM tasks \citep{hendrycks2020measuring} to gauge the efficacy of our method on reasoning in highly specialized domains. We explain below our experimental setup, result, and analysis of applying \name{} on the MMLU high-school Physics and Chemistry, and GSM8K benchmarks.

\subsection{\name{}}
Questions in the MMLU benchmarks require deeper reasoning. Furthermore, they also require understanding and application of formulae which are often physics and chemistry principles and concepts. In this case, we first demonstrate to the model abstraction skills in the form of concepts and first principles such as \textit{Newton's first law of motion}, \textit{Doppler effect}, and \textit{Gibbs free energy} etc. The implicit step-back question here is ``\textit{what are the physics or chemistry principles and concepts involved in solving this task?}''. We provide demonstrations to the model to recite the relevant principles for solving the task from its own knowledge (see Appendix~\ref{app:fewmmlu} for few-shot exemplars).

\begin{wrapfigure}{R}{0.5\textwidth}
\vspace{-6mm}
\centerline{\includegraphics[width=0.5\textwidth]{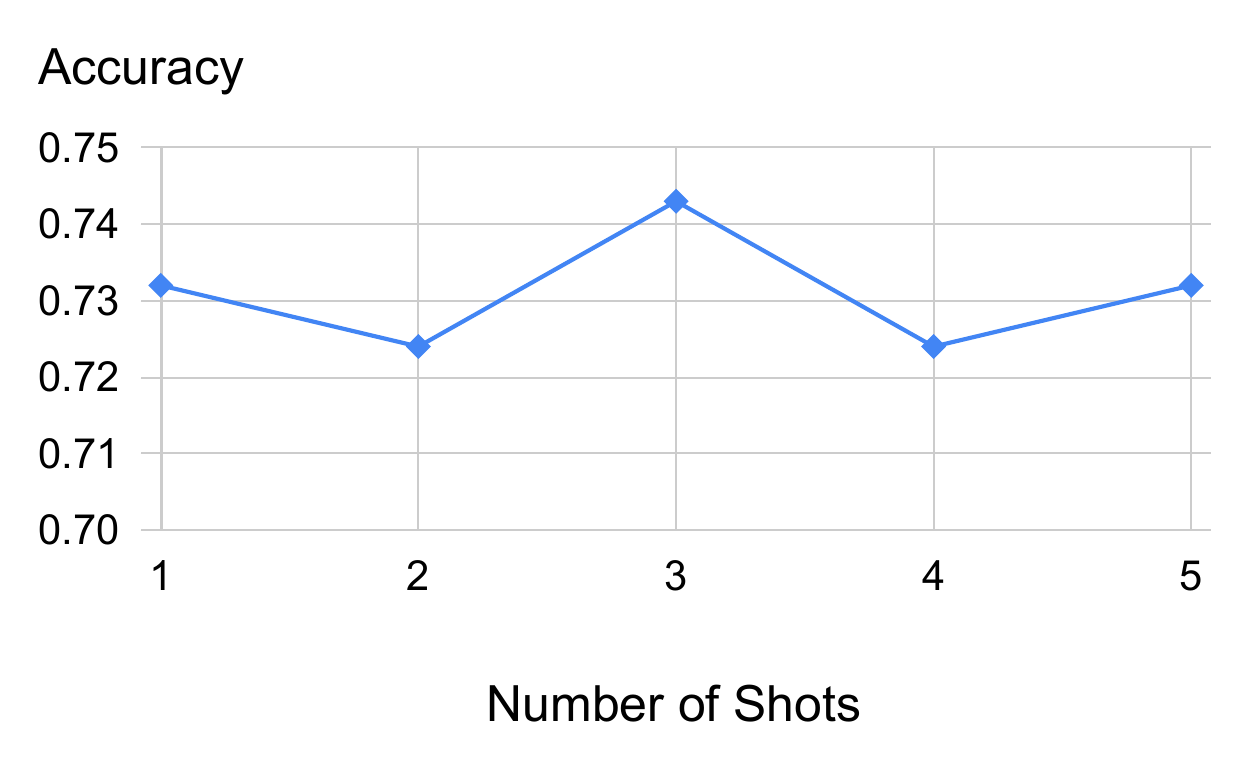}}
\caption{Ablation study of \name{} accuracy using PaLM-2L on MMLU high-school Physics against the number of few shot exemplars: robust performance with respect to a varying number of shots. }
\label{fig:mmlu_shot_abalation}
\end{wrapfigure}

\subsection{Results}
Table ~\ref{table:mmlu_physics} illustrates model performance across various setups across three model families: PaLM-2L, GPT-4, and Llama2-70B. Average accuracy over 5 evaluation runs is reported along with standard deviations (in the parentheses). PaLM-2L baseline performance is $66.4\%$ and 70.9\% on Physics and Chemistry, respectively. We find that CoT and TDB zero-shot prompting do not significantly increase model performance, which could be due to the inherent difficulty and deep reasoning associated with these tasks. PaLM-2L 1-shot and PaLM-2L + CoT 1-shot do not improve against the baseline much either, highlighting the challenge of demonstrating the reasoning steps to the model. In contrast, \name{} significantly improves model performance: +7\% and +11\% compared to PaLM-2L. Similarly, with GPT-4 and Llama2-70B models, \name{} is very competitive among all the baseline methods we tested, showing that \name{} is model-agnostic. We present the results of GSM8K in Appendix~\ref{app:gsm8k}.

\subsection{Ablation and Analysis}
\label{sec:stem_analysis}
\textbf{Few-shot Ablation}: First, in Figure~\ref{fig:mmlu_shot_abalation}, we observe that \name{} is robust to the number of few-shot exemplars of (question, principles) pairs used as demonstrations. Adding more demonstration examples beyond a single example does not lead to further improvements. This indicates that the task of retrieving the relevant principles and concepts is relatively easy through in-context learning and a single demonstration suffices. Therefore, we use a single exemplar for few-shot prompting throughout the paper except the ablation studies.

\textbf{Error Analysis}: Comparing the predictions of \name{} to the baseline PaLM-2L model for MMLU high-school Physics: we find that \name{} corrects $20.5\%$ errors from the baseline while introducing $11.9\%$ errors. 

To further understand where the errors come from in \name{}, we annotate all the wrong predictions of \name{} in the test set, and categorize them into 5 classes (see Appendix~\ref{app:mmlu_errors} for examples in each class):
\begin{itemize}[leftmargin=*, nolistsep]
\setlength{\itemsep}{1mm}
    \item \textbf{Principle Error}: The error happens at the step of Abstraction, where the first principles generated by models are wrong or incomplete.
    \item \textbf{Factual Error}: There is at least one factual error when the model recites its own factual knowledge
    \item \textbf{Math Error}: There is at least one math error in the intermediate steps when math calculations are involved in deriving the final answer.
    \item \textbf{Context Loss}: There is at least one error where the model response loses context from the question, and deviates from addressing the original question
    \item \textbf{Reasoning Error}: We define Reasoning Error as when the model makes at least one error in the intermediate Reasoning steps before arriving at the final answer.
\end{itemize}

\begin{wrapfigure}{R}{0.5\textwidth}
\vspace{-6mm}
\centerline{\includegraphics[width=0.5\textwidth]{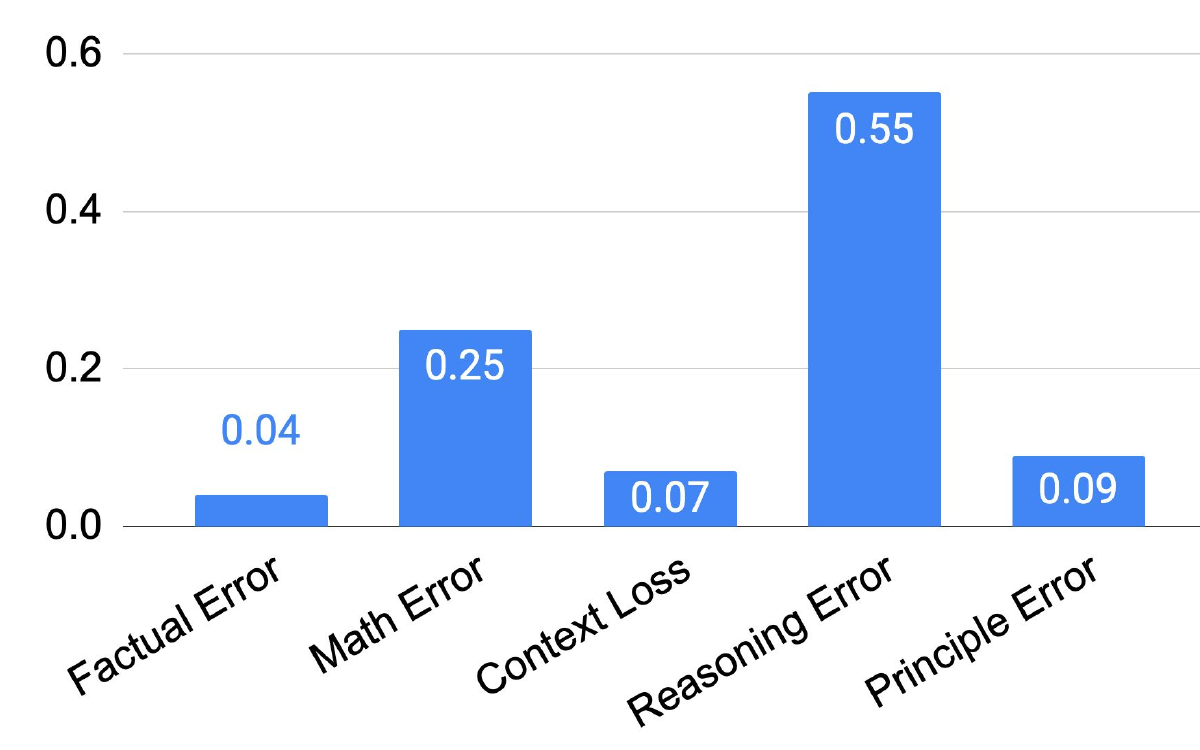}}
\caption{Error Analysis of \name{} on MMLU high-school Physics: five classes of errors Step-Back makes with Reasoning being the dominating class. }
\label{fig:mmlu_ea}
\end{wrapfigure}

All five types of errors are happening during the Reasoning step except \textit{Principle Error} which points to the failure of the Abstraction step.
As shown in Figure~\ref{fig:mmlu_ea} (right), \textit{Principle Error} comprises only a small fraction of the errors the model makes: more than $90\%$ of the errors happen at the Reasoning step. Among the four error types during Reasoning, \textit{Reasoning Error} and \textit{Math Error} are the major error categories. This corroborates with the finding in the ablation study above that very few exemplars are needed to demonstrate to LLMs the Abstraction skill. Reasoning step is still the bottleneck of how well \name{} can perform tasks such as MMLU requiring complex reasoning. For MMLU Physics specifically, the Reasoning and Math skills are critical for solving the problems successfully: even if the first principles are retrieved correctly, deep reasoning and math are involved to derive a correct final answer through a typical multi-step reasoning process.

\begin{table}[tp]
\vspace{-2mm}
\caption{Strong performance of \name{} on Knowledge QA tasks. CoT: Chain of Thought prompting, TDB: Take a Deep Breath prompting, RAG: retrieval-augmented generation. \name{} results in significant performance improvements.}
\label{table:cbqa}
\begin{center}
\begin{tabular}{l|c|cc|c}
\toprule
 Method & TimeQA & TQA Easy & TQA Hard & SituatedQA \\
\midrule 
 PaLM-2L & 41.5\% & 42.6\% & 40.4\% & 54.3\% (0.3\%) \\
 PaLM-2L 1-shot & 40.7\% & 41.7\% & 39.1\% & 51.8\% (0.6\%) \\
 PaLM-2L + CoT & 40.8\% & 41.8\% & 39.8\% & 56.4\% (0.2\%)\\
 PaLM-2L + CoT 1-shot & 38.1\% & 39.3\% & 36.8\% & 54\% (0.8\%) \\
 PaLM-2L + TDB & 40.9\% & 42.6\% & 39.1\% & 54\% (0.5\%) \\
 PaLM-2L + RAG & 57.4\% & 67.8\% &  46.8\% & 59.3\% (0.4\%) \\
  PaLM-2L + Step-Back (ours) & 66\% & 70.4\% & 61.6\% & 57.5\% (0.3\%)\\
 PaLM-2L + Step-Back + RAG (ours) & \textbf{68.7\%} & \textbf{75.2\%} & \textbf{62.3\%} & 61\% (0.4\%) \\
 \midrule 
GPT-4 & 45.6\% & 48.9\% & 42.6\% & \textbf{63.2\%} (0.4\%) \\

\bottomrule
\end{tabular}
\end{center}
\end{table}

\section{Knowledge QA}
\label{sec:timeqa}
We evaluate \name{} on question-answering benchmarks requiring intensive factual knowledge. Knowledge QA has been challenging for LLMs. In this section, we first describe the experimental setup, followed by results and analysis on \name.

\subsection{\name{}}
We evaluate \name{} on TimeQA~\citep{chen2021dataset} and SituatedQA~\citep{zhang2021situatedqa} in the Knowledge QA category. We first show the LLMs how to do Abstraction through in-context demonstrations. The step-back question ``\textit{What was Estella Leopold's education history}'' in Figure~\ref{fig:sbt_diagram} is generated by the LLM through few-shot demonstrations (see Appendix~\ref{app:fewtimeqa} for details). Given the knowledge-intensive nature of these queries, we use retrieval augmentation (RAG) in combination with \name{}. The step-back question is used to retrieve relevant facts, which work as additional context (see Table~\ref{tab:timeqa_final_prompt} for the prompt) to ground the final reasoning step.

\subsection{Results}\label{subsec:timeqaresults}
We evaluate the models on the test set of TimeQA. As shown in Table~\ref{table:cbqa}, the baseline models of GPT-4 and PaLM-2L achieved $45.6\%$ and $41.5\%$, highlighting the difficulty of the task. Applying either CoT or TDB zero-shot (and one-shot) prompting to the baseline model shows no improvement. In contrast, augmenting the baseline model by regular retrieval augmentation (RAG) improves the accuracy to $57.4\%$, highlighting the fact-intensive nature of the task. The result of Step-Back + RAG shows the effectiveness of going back to a high-level concept, which enables much more reliable retrieval augmentation: the accuracy on TimeQA achieves a remarkable $68.7\%$. 

Next, we segment TimeQA into the Easy and Hard difficulty levels provided in the original dataset. As expected, all methods perform worse on the Hard subset. While RAG can improve the Easy accuracy from $42.6\%$ to $67.8\%$, the improvement is much smaller on the Hard accuracy: $40.4\%$ to $46.8\%$. This is where \name{} shines by retrieving facts regarding high-level concepts to ground the final reasoning: Step-Back + RAG further improves the Hard accuracy to $62.3\%$, outperforming GPT-4's $42.6\%$ from GPT-4. We hypothesize that facts regarding the high-level concepts (such as \textit{education history}) are much more accessible than the low-level details.

On the SituatedQA benchmark, we observe a moderate quality gain from $54.3\%$ to our best method of Step-Back + RAG ($61\%$) with a small gap to GPT-4's $63.2\%$. Similar to TimeQA, prompting techniques such as CoT and TDB don't help significantly for SituatedQA.   

\begin{figure*}[!h]
\centerline{\includegraphics[width=0.9\linewidth]{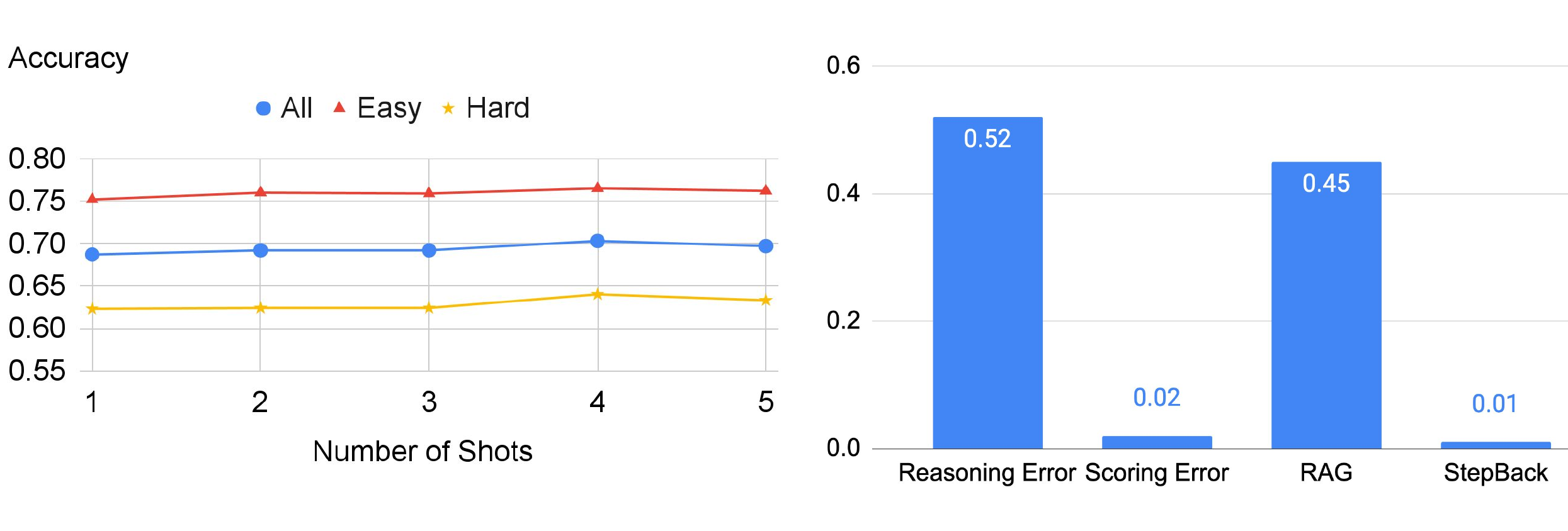}}
\caption{Ablation and error analysis of \name{} on TimeQA. \textit{Left}: ablation against the number of few-shot exemplars. \textit{Right}: four classes of errors Step-Back makes with Reasoning and RAG being the dominant error sources.
}
\label{fig:timeqa_error_analysis}
\end{figure*}

\subsection{Ablation and Analysis}
\label{sec:timeqa_analysis}
\textbf{Few-shot Ablation}: We observe in Figure~\ref{fig:timeqa_error_analysis} (left) that the performance of \name{} on TimeQA is robust to the number of exemplars used in demonstration, highlighting again the sample efficiency of in-context learning Abstraction skills for models like PaLM-2L.

\textbf{Error Analysis:} Figure~\ref{fig:timeqa_error_analysis} (right) shows the breakdown of all the remaining errors made by \name{} on TimeQA. Similar to Section~\ref{sec:stem_analysis}, we categorize the errors into
\begin{itemize}[leftmargin=*, nolistsep]
\setlength{\itemsep}{1mm}
    \item \textbf{StepBack}: The step-back question generated is not helpful in solving the task.
    \item \textbf{RAG}: RAG fails to retrieve relevant information despite that the step-back question is on target.
    \item \textbf{Scoring Error}: The evaluation by the judge model made a mistake.
    \item \textbf{Reasoning Error}: The retrieved context is relevant, but the model still fails to reason through the context to arrive at the right answer.
\end{itemize}

We find that the StepBack rarely fails. In contrast, we find more than half of the errors are due to reasoning errors. Additionally, $45\%$ of errors are due to failure in retrieving the right information despite that Abstraction provided by step-back makes it a much easier task. This reflects the difficulty level of the TimeQA task.  
Additional error analysis of TimeQA is in Appendix~\ref{app:timeqaerrora}.

\begin{table}%
\caption{Results of \name{} on Multi-Hop Reasoning. CoT: Chain of Thought prompting, TDB: Take a Deep Breath prompting, RAG: retrieval augmentation generation. The average accuracy is over 5 evaluation runs with the standard deviations included in the parentheses.}
\label{table:strategyqa}
\begin{center}
\begin{tabular}{ll|c|c}
\toprule
 & Method & MuSiQue & StrategyQA\\

\midrule 
 & PaLM-2L  & 35.5\% (3\%) & 82.8\% (0.7\%)\\
 & PaLM-2L 1-shot  & 29.0\% (0.5\%)& 76.6\% (0.5\%)\\
 & PaLM-2L + CoT  & 38.7\% (3.2\%) & 83.6\% (0.4\%)\\
 & PaLM-2L + CoT 1-shot & 38.5\% (2.2\%) & 76.8\% (1.4\%) \\
 & PaLM-2L + TDB & 39.0\% (2.3\%) &82.7\% (0.9\%) \\
 & PaLM-2L + RAG & 39.6\% (2.8\%) & 84.2\% (0.5\%)  \\
 & PaLM-2L + Step-Back (ours)& 42.6\% (3.1\%) & 82.7\% (0.4\%) \\
 & PaLM-2L + Step-Back + RAG (ours) & \textbf{42.8}\% (2.0\%)& \textbf{86.4\%} (1\%)  \\
 \midrule 
 & GPT-4 & 38.5\% (0.2\%)& 78.3\% (1.1\%) \\
\bottomrule
\end{tabular}
\end{center}
\end{table}

\section{Multi-Hop Reasoning}
\label{sec:multihop_reasoning}

We evaluate \name{} on challenging Multi-Hop reasoning benchmark MuSiQue \citep{trivedi2022musique} and StrategyQA \citep{geva2021did}. We follow the same protocol as Section~\ref{sec:timeqa} to implement \name{}.

Table ~\ref{table:strategyqa} shows performance of various baselines on the dev set of MuSiQue and StrategyQA. Baseline performance of PaLM-2L and GPT-4 are low ($35.5\%$ and $38.5\%$ for PaLM-2L and GPT-4 respectively) in MuSiQue since it is a hard multihop reasoning benchmark. In contrast, StrategyQA has stronger baselines ($82.8\%$ and $78.3\%$ for PaLM-2L and GPT-4 respectively) probably because it is a binary classification task. CoT and TDB improve model performance a bit in the case of MuSiQue ($\sim$ 3\%\ and 3.5\% respectively) which can be attributed to the inherent reasoning nature of this task where these methods are shown to be helpful. In the case of StrategyQA, there is no significant performance gain with CoT and TDB which could be due to the high baseline performance in this task, with limited scope for these prompting methods to improve performance. Often, 1-shot performance is significantly lower than their zero-shot methods, which could be attributed to potential example bias~\citep{zhao2021calibrate, parmar2023don}. RAG improves model performance ($\sim$ 4\%\  and 2\% for MuSiQue and StrategyQA respectively.) \name{} with the power of abstraction produces the best performance of all methods: $42.8\%$ in MuSiQue and $86.4\%$ in StrategyQA, significantly outperforming GPT-4 on both tasks. We present a detailed error analysis on StrategyQA in Appendix~\ref{app:startegyqaerrora}.

\section{Discussion}
Abstraction helps humans to solve complex tasks by removing irrelevant details and distilling high-level concepts and principles to guide the problem-solving process. 
\name{} breaks complex tasks such as knowledge-intensive QA, multi-hop reasoning, and science questions into two separate steps of Abstraction and Reasoning. We demonstrate through empirical experiments that Abstraction is an easy skill for the LLMs such as PaLM-2L via sample-efficient in-context learning. Grounding on the high-level concepts and principles, LLMs can leverage their intrinsic Reasoning capabilities to derive the solution. This reduces the chance of reasoning failures in the intermediate steps and is shown to improve the performance on a wide range of complex reasoning tasks. Despite the success, through error analysis, we find that Reasoning is still one of the hardest skills for LLMs to acquire: it is still the dominant failure mode even after the large reduction of task complexity by \name{}.

Nevertheless, Abstraction is neither necessary nor possible in all scenarios. For instance, the task can be as simple as \textit{who was the president of the United States in 2000?}, in which case there is no such need to step back and ask a high-level question as the answer to such questions is readily available. Questions such as \textit{what is the speed of light?} point to the first principles themselves. Doing Abstraction in this case would not make a difference either. 

\section{Related Work}

\subsection{Prompting}
Few-shot prompting~\citep{brown2020language, liu2023pre, mishra2022reframing, wei2022chain} has significantly improved model performance across a range of tasks without requiring updating any model parameters. Our work \name{} is in the same category as the chain-of-thought prompting~~\citep{wei2022chain} and scratchpad~\citep{nye2021show} owing to its simplicity and generic nature. But our approach is focused on the key idea of abstraction which is inspired from the fact that taking a step back often helps humans in performing complex tasks. Our work is also related to the recitation-augmented language models~\citep{sun2022recitation}; however in contrast to their work, we explicitly perform step-back and abstraction, with optional use of retrieval augmentation depending on the nature of the task at hand.

\subsection{Decomposition}
Decomposing a task into simpler tasks and solving these tasks to complete the original task has been an effective way~\citep{zhou2022least, patel2022question, khot2022decomposed,  press2022measuring} to improve model performance on complex tasks. Several prompting methods have been successful in this regard. Our work \name{}, in contrast, is on making the question more abstract and high-level, which is different from decomposition that is often a low-level breakdowns of the original question. For instance, a generic question for \textit{which employer did Steve Jobs work for in 1990?} could be \textit{what is the employment history of Steve Jobs?} While decomposition would lead to sub-questions such as \textit{What was Steve Jobs doing in 1990?}, \textit{Was Steve Jobs employed in 1990?} and \textit{If Steve Jobs was employed, who was his employer?} Furthermore, abstract questions such as \textit{what is the employment history of Steve Jobs?} are often generic in nature to have a many-to-one mapping since many questions (e.g. \textit{which employer did Steve Jobs work for in 1990?} and \textit{which employer did Steve Jobs work for in 2000?}) can have the same abstract question. This is in contrast to decomposition where there is often a one-to-many mapping since there are multiple decomposed sub-problems necessary to solve a given question.

\section{Conclusion}
We introduce \name{} as a simple yet generic method to elicit deep reasoning via abstraction in large language models. Experimentation on LLMs across fact-seeking, commonsense reasoning and domain-specific reasoning benchmarks shows that \name{} significantly improves model performance. We hypothesize that abstraction helps models to hallucinate less and reason better, probably reflecting the true nature of the model which are often hidden while responding to the original question without abstraction. We hope our work will inspire more human-inspired approaches to elicit the hidden potential of large language models.

\section*{Acknowledgement} We thank Andrew Dai,  Adams Yu and Hannah Rashkin of Google DeepMind for their insightful feedback on this paper.

\bibliography{iclr2024_conference}
\bibliographystyle{iclr2024_conference}

\clearpage

\appendix

\section{GSM8K Results, and Error Analysis}\label{app:timeqaerrora}
\subsection{GSM8K Results}\label{app:gsm8k}
We present in Table~\ref{table:gsm8k} the results of \name{} on GSM8K along with other strong baselines from PaLM-2L runs. We observe that \name{} achieved competitive performance together with zero-shot CoT and 1-shot standard prompting. We hypothesize that the simplicity of principles (e.g. addition, subtraction, etc.) in GSM8K makes it not absolutely necessary to retrieve the principles first before reasoning. Nonetheless, we still find that \name{} is the most competitive among all the prompting methods we tested, including the ``Take a Deep Breath'' prompting optimized for GSM8K in \cite{yang2023large} and Decomposed Prompting in ~\cite{khot2022decomposed}.

\begin{table}
\caption{Step-Back Prompting on GSM8K. CoT: zero-shot Chain of Thought prompting \citep{kojima2022large}, TDB: Take a Deep Breath prompting \citep{yang2023large}, DP: Decomposed Prompting \citep{khot2022decomposed}. The Table reports the average accuracy over 5 evaluation runs, with standard deviations in the parentheses.}
\label{table:gsm8k}
\begin{center}
\begin{tabular}{l|c}
\toprule
Method & GSM8K \\
\midrule 
PaLM-2L & 75.8\% (0.2\%)\\
PaLM-2L 1-shot & \textbf{84.5\%} (0.4\%)\\
PaLM-2L + CoT & \textbf{84.4\%} (0.2\%)\\
PaLM-2L + CoT 1-shot & 81\% (0.2\%)\\ 
PaLM-2L + TDB & 82.2\% (0.2\%)\\
PaLM-2L + DP & 82.2\% (0.08\%)\\
\midrule 
PaLM-2L + Step-Back (ours) & \textbf{84.3\%} (0.2\%)\\
\bottomrule
\end{tabular}
\end{center}
\end{table}

\subsection{TimeQA Error Analysis} 
\begin{figure}[!h]
\centerline{\includegraphics[width=\linewidth]{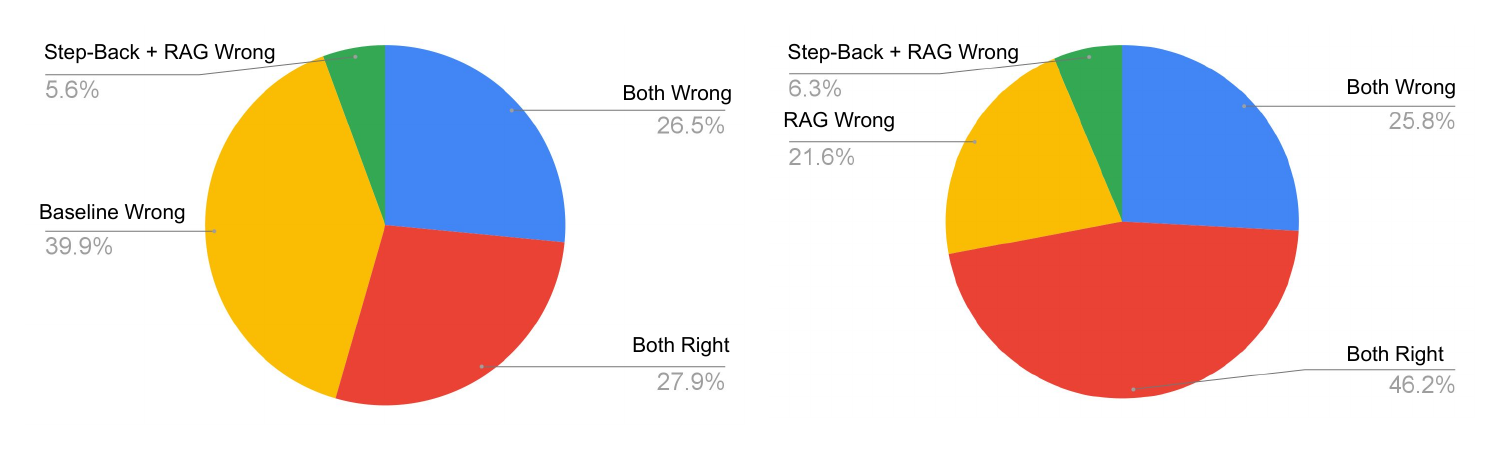}}
\caption{Error Analysis of Step-Back Prompting on TimeQA. \textit{Left}: Step-Back + RAG vs Baseline predictions. \textit{Right}: Step-Back RAG vs RAG predictions. Step-Back + RAG can fix $39.9\%$ of the predictions where the baseline prediction is wrong while causing $5.6\%$ errors. Furthermore, Step-Back + RAG fixes $21.6\%$ errors coming from RAG. The $\%$ of errors introduced by \name{} to RAG is still relatively low ($6.3\%$).}
\label{fig:figure_timeqa_ea}
\end{figure}

We conduct error analysis to understand where \name{} fixes the errors the baseline models make.
Figure~\ref{fig:figure_timeqa_ea} shows that compared to the predictions of baseline PaLM-2L, \name{} can fix $39.9\%$ of the predictions where the baseline prediction is wrong, while causing $5.6\%$ errors.Furthermore, Step-Back + RAG fixes $21.6\%$ errors coming from RAG. The $\%$ of errors introduced by \name{} to RAG is still relatively low ($6.3\%$). Together, this shows that the \name{} is helpful most of the time, signifying the need and effectiveness of doing Abstraction before directly addressing the original question.
\subsection{StrategyQA Error Analysis} \label{app:startegyqaerrora}
\begin{figure}[b]
\centerline{\includegraphics[width=\linewidth]{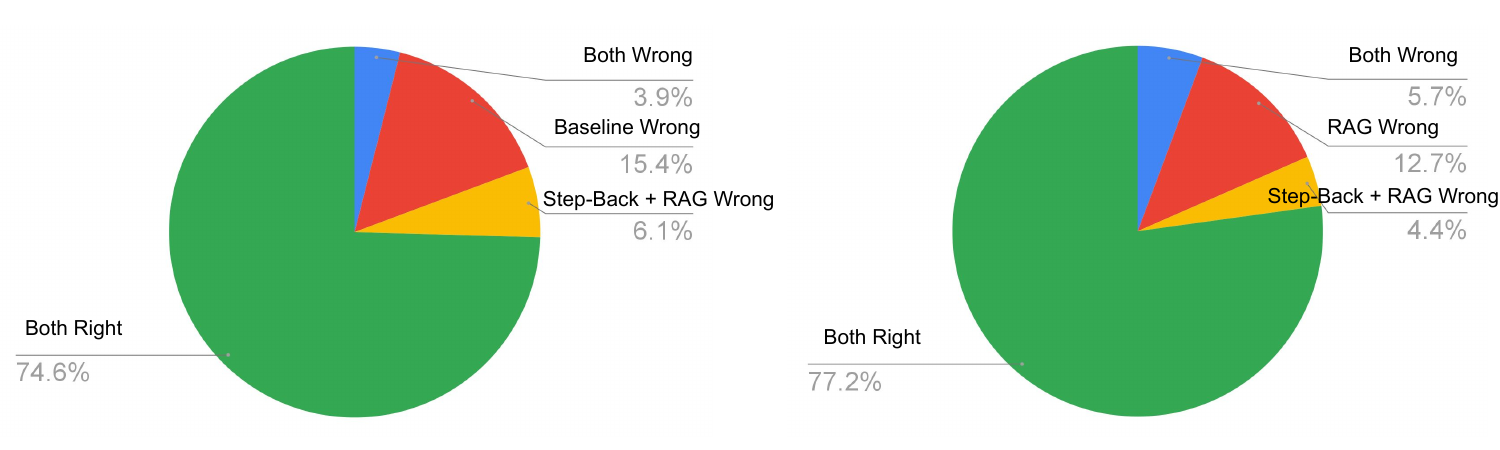}}
\caption{Error Analysis of Step-Back Prompting on StrategyQA. \textit{Left}: Step-Back + RAG vs Baseline predictions. \textit{Right}: Step-Back + RAG vs RAG predictions. Step-Back + RAG is able to turn $15.4\%$ wrong predictions into correct predictions, while leading to $6.1\%$ errors the other way around. 
Furthermore, Step-Back + RAG fixes $12.7\%$ errors coming from RAG. The errors introduced to RAG by Step-Back are just $4.4\%$.}
\label{fig:figure4_sqa_ea}
\end{figure}

Figure~\ref{fig:figure4_sqa_ea} shows the error analysis of StrategyQA on the predictions of Step-Back + RAG against the baseline model and the raw retrieval augmentation variant of PaLM-2L. Compared to the baseline, Step-Back + RAG can turn $15.4\%$ wrong predictions into correct predictions, while leading to $6.1\%$ errors the other way around. 
Furthermore, Step-Back + RAG fixes $12.7\%$ errors coming from RAG. The errors introduced to RAG by Step-Back are just $4.4\%$.

\section{Dataset Details} 
\label{app:dataset_details}
Table~\ref{app:eval_stats} shows the split and number of examples used for evaluations in TimeQA, StrategyQA, MMLU, and GSM8K.

\begin{table}%
\caption{Stats of the evaluation datasets used in this paper.}
\label{app:eval_stats}
\begin{center}
\begin{tabular}{l|c|c|c}
\toprule
Domain &Dataset & Split & Number of Examples \\
\midrule 
\midrule 
STEM & MMLU high-school Physics & Test & 151 \\
& MMLU high-school Chemistry & Test & 203 \\
& GSM8K & Test & 1319 \\ 
\midrule 
Knowledge QA & TimeQA & Test & 5226 \\
& TimeQA Easy & Test & 2613 \\
& TimeQA Hard & Test & 2613 \\
& SituatedQA & Test & 2901 \\
\midrule 
Multi-hop Reasoning & MuSiQue & Dev & 2417 \\
& StrategyQA & Dev & 229 \\
\bottomrule
\end{tabular}
\end{center}
\end{table}

\section{Evaluation Details}\label{app:eval}
\subsection{Few-shot Examples for Evaluation with PaLM-2L}\label{app:fewshotexamples}
Given the model free-form outputs and the target label, we use one positive and one negative output as few-shot examples to demonstrate to the scoring model how to score the output. Table ~\ref{tab:fewevalpalm} illustrates the prompt we used for the scoring model. We parse out the ``Yes" or ``No" answer from the scoring model output as a TRUE or FALSE score of the model output.

\begin{table*}
    \centering
    \begin{tabular}{p{10cm}}
        \toprule
            Are the following two answers to the given question equivalent? Do not consider whether the answers are right or wrong, but only whether they are equivalent. Directly state "Yes" or "No".\newline 
            \textbf{Question}: Which title was conferred to Anna Muzychuk in 2007?\newline 
            \textbf{Answer 1}: Anna Muzychuk was conferred the title of International Master (IM) in 2007. She earned the title by scoring three norms in rapid chess tournaments.\newline 
            \textbf{Answer 2}: International Master \newline 
            \textbf{Answer 1 (short)}: International Master\newline 
            \textbf{Answer 2 (short)}: International Master\newline 
            \textbf{Are the two answers equivalent?} Yes \newline 
            \textbf{Question}: What state is Seattle located in?\newline 
            \textbf{Answer 1}: Seattle is in Washington State.\newline 
            \textbf{Answer 2}: The answer is George Washington.\newline 
            \textbf{Answer 1 (short)}: Washington State\newline 
            \textbf{Answer 2 (short)}: George Washington\newline 
            \textbf{Are the two answers equivalent?} No\newline
            \textbf{Question}: $<$Question$>$\newline
            \textbf{Answer 1}: $<$Model Output$>$\newline
            \textbf{Answer 2}: $<$Target Label$>$\\
        \bottomrule
        \end{tabular}
        \caption{Illustration of few shot evaluation with the PaLM-2L model.
        }
        \label{tab:fewevalpalm}
\end{table*}

\subsection{Hyper-parameters for Evaluation with PaLM-2L}\label{app:fewshoteval}
We use PaLM-2L as the scoring model for evaluation. We experiment with different sampling temperatures, and find that $T=1$ gives us a highly-accurate evaluation. For example, we sampled $100$ test examples and the model predictions, and manually rated the correctness of the model scoring. We found that out of 4 trials, the model scoring agrees with human ratings $97\%$, $98\%$, $99\%$ and $99\%$ of the time. 

\section{Prompts and Few shot Examples}
\subsection{STEM}
\label{app:fewmmlu}
For MMLU high-school Physics and Chemistry, we first prompt the model to generate the first principles behind the question. Using the generated first principles, we further prompt the model to generate the final answer through few-shot demonstrations
The prompt generating first principles is shown in Table~\ref{tab:mmlu_principle_prompt} for MMLU high-school Physics and Chemistry.

\begin{table*}
    \centering
    \begin{tabular}{p{10cm}}
        \toprule
         MMLU Physics/Chemistry First-Principle Prompt \\
      \midrule
            You are an expert at Physics/Chemistry. You are given a Physics/Chemistry problem. Your task is to extract the Physics/Chemistry concepts and principles involved in solving the problem. Here are a few examples:\newline \newline
            Question: $<$Question Example1$>$ \newline
            Principles Involved: $<$Principles Example1$>$ \newline
            ... \newline
            Question: $<$Question Example5$>$ \newline
            Principles Involved: $<$Principles Example5$>$ \newline
            Question: $<$Question$>$ \newline
            Principles Involved:
            \\
        \bottomrule
        \end{tabular}
        \caption{Prompt of extracting the underlying principles involved in MMLU physics and chemistry questions.}
        \label{tab:mmlu_principle_prompt}
\end{table*}

After extracting the first principles of solving a particular question, we formulate the prompt in Table~\ref{tab:mmlu_answer_prompt} to query the model for the final answer.

\begin{table*}
    \centering
    \begin{tabular}{p{10cm}}
        \toprule
         MMLU Physics/Chemistry Final Answer Prompt \\
      \midrule
            You are an expert at Physics/Chemistry. You are given a Physics/Chemistry problem and a set of principles involved in solving the problem. Solve the problem step by step by following the principles. Here are a few examples:\newline \newline
            Question: $<$Question Example1$>$ \newline
            Principles: $<$Principles Example1$>$ \newline
            Answer: $<$Answer Example1$>$ \newline
            ...\newline
            Question: $<$Question Example5$>$ \newline
            Principles: $<$Principles Example5$>$ \newline
            Answer: $<$Answer Example5$>$ \newline
            Question: $<$Question$>$ \newline
            Principles: $<$Principles$>$ \newline
            Answer:
            \\
        \bottomrule
        \end{tabular}
        \caption{Prompt of querying the model for final answer with first principles behind the question in MMLU high-school Physics and Chemistry.}
        \label{tab:mmlu_answer_prompt}
\end{table*}

Tables~\ref{tab:mmlu_stepback_exemplars_1}-\ref{tab:mmlu_stepback_exemplars_chem_1} show one demonstration exemplar of Question-Principles-Answer triplets for MMLU high-school Physics and Chemistry, respectively. For GSM8K, given the simplicity of the principles, we directly combine the principles and the solution in the demonstration exemplar. Table~\ref{tab:gsm8k_stepback_prompt} shows the exemplar we used in the paper.
\begin{table*}
    \centering
    \begin{tabular}{p{2cm}p{11cm}}
        \toprule
         Question & A spherical conductor carries a net charge. How is this charge distributed on the sphere?\newline
            0) The charge is evenly distributed on the surface.\newline
            1) The charge resides on the surface only; the distribution of charge on the surface depends on what other charged objects are near the sphere.\newline
            2) The charge moves continually within the sphere.\newline
            3) The charge is distributed uniformly throughout the sphere.\\
      \toprule
      Principles & Coulomb's Law: the force between two charged particles is proportional to the product of their charges and inversely proportional to the square of the distance between them, F = k * q1 * q2 / r, where F is the electric force, k is a constant, q1 and q2 are the charges the particles carry, and r is the distance between them. \\
      \toprule
      Answer & Using the Principles of Coulomb's Law, we can solve the problem as following: \newline
            Step 1: Apply Coulomb's Law to find out how charges are distributed on the surface.\newline
            In the case of a spherical conductor, the charges on the surface will repel each other. The further apart the charges are, the less force they will exert on each other. Therefore, the charges will distribute themselves evenly on the surface of the sphere, as this is the configuration that minimizes the repulsive force between them.\newline
            
            Step 2: Apply Coulomb's Law to find out what happens if there are other charges present.\newline
            The distribution of charge on the surface may also be affected by the presence of other charged objects near the sphere. For example, if a negatively charged object is brought near a positively charged sphere, the negative charges on the sphere will be repelled and will move to the opposite side of the sphere. This will result in a non-uniform distribution of charge on the surface of the sphere.\newline
            
            Therefore, the correct answer is 1) The charge resides on the surface only; the distribution of charge on the surface depends on what other charged objects are near the sphere. \\
        \bottomrule
        \end{tabular}
        \caption{Few-shot demonstration exemplar for MMLU high-school Physics.
        }
        \label{tab:mmlu_stepback_exemplars_1}
\end{table*}

\begin{table*}
    \centering
    \begin{tabular}{p{2cm}p{11cm}}
        \toprule
         Question & A sample of an unknown chloride compound was dissolved in water, and then titrated with excess Pb(NO3)2 to create a precipitate. After drying, it is determined there are 0.0050 mol of precipitate present. What mass of chloride is present in the original sample?\newline
         0) 0.177 g\newline
         1) 0.355 g\newline
         2) 0.522 g\newline
         3) 0.710 g\\
      \toprule
      Principles & Precipitation reactions: Precipitation reactions occur when two soluble salts are mixed and form an insoluble product, called a precipitate. The precipitate can be separated from the solution by filtration or centrifugation.\newline
      Molar mass: The molar mass of a substance is the mass of one mole of that substance. The molar mass is expressed in grams per mole (g/mol).\newline
      Limiting reactant: The limiting reactant is the reactant that is completely consumed in a chemical reaction. The amount of product formed is determined by the amount of limiting reactant.\\
      \toprule
      Answer & Assuming the unknown chloride compound is MCl, where M represents the metal cation, the balanced chemical equation for the precipitation reaction is:\newline
      $Pb(NO_3)_2(aq)$ + $2MCl(aq) \longrightarrow PbCl_2(s)$ + $2MNO_3(aq)$ \newline
      
      Since $Pb(NO_3)_2$ is in excess, MCl is the limiting reactant. The stoichiometry of the reaction indicates that 2 moles of MCl produce 1 mole of $PbCl_2$ precipitate. Therefore, 0.0050 mol of $PbCl_2$ corresponds to 0.010 mol of MCl.\newline
      
      The mass of chloride in the original sample can be calculated using the molar mass of chloride (35.45 g/mol):\newline
      0.010 mol Cl × 35.45 g/mol = 0.355 g Cl \newline
      
      The correct answer is 1) 0.355 g.\\
        \bottomrule
        \end{tabular}
        \caption{Few-shot demonstration exemplar for MMLU high-school Chemistry.
        }
        \label{tab:mmlu_stepback_exemplars_chem_1}
\end{table*}

\begin{table*}
    \centering
    \begin{tabular}{p{2cm}p{11cm}}
        \toprule
         Question & Bella bought stamps at the post office. Some of the stamps had a snowflake design, some had a truck design, and some had a rose design. Bella bought 11 snowflake stamps. She bought 9 more truck stamps than snowflake stamps, and 13 fewer rose stamps than truck stamps. How many stamps did Bella buy in all? \\
      \toprule 
      Answer & Let us find and apply the math principles to solve the problem step by step:\newline
      
      Step 1. Addition: Calculate the number of truck stamps.\newline
      Bella bought 11 snowflake stamps. She bought 9 more truck stamps than snowflake stamps: there are 11 + 9 = 20 truck stamps.\newline
      
      Step 2. Subtraction: Calculate the number of rose stamps.\newline
      Bella bought 13 fewer rose stamps than truck stamps: there are 20 - 13 = 7 rose stamps.\newline
      
      Step 3. Addition: Calculate the total number of stamps in all three colors.\newline
      Bella bought 11 snowflake stamps, 20 truck stamps, 7 rose stamps: there are 11 + 20 + 7 = 38 stamps in total.\newline
      
      Conclusion: Bella bought 38 stamps in all.\\
        \bottomrule
        \end{tabular}
        \caption{Step-Back few-shot demonstration exemplar for GSM8K.
        }
        \label{tab:gsm8k_stepback_prompt}
\end{table*}

\subsection{Knowledge QA}
\label{app:fewtimeqa}
We use the following prompting in Table~\ref{tab:timeqa_stepback_prompt} to demonstrate to the LLM on asking a step-back question for TimeQA and SituatedQA including up to 5 exemplar demonstrations of pairs of Original Question and Step-back Question.
\begin{table*}
    \centering
    \begin{tabular}{p{10cm}}
        \toprule
         Knowledge QA Step-Back Prompt \\
      \midrule
            You are an expert at world knowledge. Your task is to step back and paraphrase a question to a more generic step-back question, which is easier to answer. Here are a few examples: \newline \newline
            Original Question: $<$Original Question Example1$>$ \newline
            Stepback Question: $<$Stepback Question Example1$>$ \newline
            ... \newline
            Original Question: $<$Original Question Example5$>$ \newline
            Stepback Question: $<$Stepback Question Example5$>$ \newline
            Original Question: $<$Original Question$>$ \newline
            Stepback Question:
            \\
        \bottomrule
        \end{tabular}
        \caption{Prompt of asking step-back question in Knowledge QA tasks.}
        \label{tab:timeqa_stepback_prompt}
\end{table*}

Table~\ref{tab:timeqa_stepback_exemplars} shows 5 exemplars from the Train split of TimeQA and SituatedQA as demonstrations of asking step-back questions.
\begin{table*}
    \centering
    \begin{tabular}{p{2cm}p{5.5cm}p{5.5cm}}
        \toprule
         dataset & Original Question & Step-back Question\\
      \midrule
            TimeQA & Which position did Knox Cunningham hold from May 1955 to Apr 1956? & Which positions have Knox Cunningham held in his career? \\
            \midrule
            TimeQA & Who was the spouse of Anna Karina from 1968 to 1974? & Who were the spouses of Anna Karina? \\
            \midrule
            TimeQA & Which team did Thierry Audel play for from 2007 to 2008? & Which teams did Thierry Audel play for in his career? \\
            \midrule
            TimeQA & What was the operator of GCR Class 11E from 1913 to Dec 1922? & What were the operators of GCR Class 11E in history? \\
            \midrule
            TimeQA & Which country did Sokolovsko belong to from 1392 to 1525? & Which countries did Sokolovsko belong to in history? \\
            \midrule
            SituatedQA & when was the last time a team from canada won the stanley cup as of 2002 & which years did a team from canada won the stanley cup as of 2002 \\
            \midrule
            SituatedQA & when did england last get to the semi final in a world cup as of 2019 & which years did england get to the semi final in a world cup as of 2019? \\
            \midrule
            SituatedQA & what is the biggest hotel in las vegas nv as of November 28, 1993 & what is the size of the hotels in las vegas nv as of November 28, 1993 \\
            \midrule
            SituatedQA & who has scored most runs in t20 matches as of 2017 & What are the runs of players in t20 matches as of 2017 \\
            \midrule
            SituatedQA & who is the highest paid player in the nba this season as of 2017 & what is the salary of the high paid players in the nba this season as of 2017 \\
            \midrule
        \bottomrule
        \end{tabular}
        \caption{Few-shot demonstration exemplars for asking step-back questions in TimeQA and SituatedQA.
        }
        \label{tab:timeqa_stepback_exemplars}
\end{table*}

The step-back question is extracted from the model output using the prompt. Using the step-back question, we do retrieval augmentation. Using both the retrieval augmentations from the original question and the step-back question, we formulate the final prompt to query the model for the final answer, as shown in Table~\ref{tab:timeqa_final_prompt}.
\begin{table*}
    \centering
    \begin{tabular}{p{10cm}}
        \toprule
         Knowledge QA Final-Answer Prompt \\
      \midrule
            You are an expert of world knowledge. I am going to ask you a question. Your response should be comprehensive and not contradicted with the following context if they are relevant. Otherwise, ignore them if they are not relevant.\newline \newline
            $<$Passage from original retrieval augmentation$>$ \newline
            $<$Passage from step-back retrieval augmentation$>$ \newline\newline
            Original Question: $<$Original Question$>$ \newline
            Answer:
            \\
        \bottomrule
        \end{tabular}
        \caption{Prompt of querying the model for final answer with additional contexts from original and step-back retrieval augmentations in TimeQA and SituatedQA}
        \label{tab:timeqa_final_prompt}
\end{table*}

\subsection{Multi-Hop Reasoning}
\label{app:fewstrategyqa}
For Multi-Hop Reasoning, we use the same prompting template as in Knowledge QA to ask the step-back question, and query for the final answer given the retrieval augmentations. Table~\ref{tab:strategyqa_stepback_exemplars} shows 5 demonstration exemplars for asking step-back questions from the Train split of MuSiQue and StrategyQA.

\begin{table*}
    \centering
    \begin{tabular}{p{2cm}p{5cm}p{5cm}}
        \toprule
         dataset & Original Question & Step-back Question\\
      \midrule
            MuSiQue & at year saw the creation of the region where the county of Hertfordshire is located? & which region is the county of Hertfordshire located? \\
            \midrule
            MuSiQue & Jan Šindel's was born in what country? & what is Jan Šindel's personal history? \\
            \midrule
            MuSiQue & When was the abolishment of the studio that distributed The Game? & which studio distributed The Game? \\
            \midrule
            MuSiQue & What city is the person who broadened the doctrine of philosophy of language from? & who broadened the doctrine of philosophy of language \\
            \midrule
            MuSiQue & When was the baseball team winning the world series in 2015 baseball created? & which baseball team won the world series in 2015 baseball? \\
            \midrule
            StrategyQA & Could the members of The Police perform lawful arrests? & what can the members of The Police do? \\
            \midrule
            StrategyQA & Would a Monoamine Oxidase candy bar cheer up a depressed friend? & What are the effects of Monoamine Oxidase? \\
            \midrule
            StrategyQA & Would a dog respond to bell before Grey seal? & Would a dog respond to bell before Grey seal? \\
            \midrule
            StrategyQA & Is shrimp scampi definitely free of plastic? & what is shrimp scampi made of? \\
            \midrule
            StrategyQA & Do the anchors on Rede Globo speak Chinese? & What languages do the anchors on Rede Globo speak? \\
            \midrule
        \bottomrule
        \end{tabular}
        \caption{Few-shot demonstration exemplars for asking step-back questions in MuSiQue and StrategyQA.
        }
        \label{tab:strategyqa_stepback_exemplars}
\end{table*}

\subsection{Baseline Prompts}
\label{app:baseline_prompt}
For standard zero-shot and few-shot prompting of the baseline model, we formulate the prompt using the template in Table~\ref{tab:baseline_answer_prompt} with up to 1 exemplars.

\begin{table*}
    \centering
    \begin{tabular}{p{10cm}}
        \toprule
         Baseline few-shot Prompt \\
      \midrule
            You are an expert of world knowledge and physics. Your task is to solve the following question. Here are a few examples:\newline \newline
            Question: $<$Question Example$>$ \newline
            Answer: $<$Answer Example$>$ \newline
            Question: $<$Question$>$ \newline
            Answer:
            \\
        \bottomrule
        \end{tabular}
        \caption{Prompt of querying the baseline model for final answer with few-shot demonstration exemplars.}
        \label{tab:baseline_answer_prompt}
\end{table*}

\begin{table*}
    \centering
    \begin{tabular}{p{1.5cm}p{5cm}p{6.5cm}}
        \toprule
         dataset & Question & Answer\\
      \midrule
            MMLU Physics & A point charge, Q1 = +1 mC, is fixed at the origin. How much work is required to move another charge, Q2 = +8 $\mu$C, from the point (0, 4 meters) to the point (3 meters, 0)? \newline 
            0) 3.5 J 1) 6.0 J 2) 22.5 J 3) 40 J & The work required to move a charge in an electric field is equal to the change in potential energy of the charge. The potential energy of a charge in an electric field is given by: U = k * Q1 * Q2 / r
            where: k is the Coulomb constant, Q1 and Q2 are the charges, r is the distance between the charges.
            In this case, the initial potential energy of Q2 is:
            U1 = k * Q1 * Q2 / 4 meters\newline
            
            The final potential energy of Q2 is:
            U2 = k * Q1 * Q2 / 3 meters\newline
            
            The change in potential energy is:  
            $\Delta$U = U2 - U1 = k * Q1 * Q2 * (1/3 meters - 1/4 meters)\newline
            
            Plugging in the values, we get:
            $\Delta$U = $(9 \times 10^9 N*m^2/C^2) \times (1 \times 10^{-3} C) * (8 \times 10^{-6} C) * (1/3m - 1/4m) = 6 J$.
            The answer is 1) 6.0 J. \\
            \midrule
            MMLU Chemistry & A sample of an unknown chloride compound was dissolved in water, and then titrated with excess Pb(NO3)2 to create a precipitate. After drying, it is determined there are 0.0050 mol of precipitate present. What mass of chloride is present in the original sample?\newline
            0) 0.177 g 1) 0.355 g 2) 0.522 g 3) 0.710 g & 0.0050 moles of lead chloride precipitate contain 0.010 moles of chloride ions due to the 2:1 mole ratio between chloride ions and lead chloride.\newline
            
            The mass of 0.010 moles of chloride ions is 0.355 g.\newline
            
            Therefore, the mass of chloride present in the original sample is 0.355 g. \\
            \midrule
            GSM8K & Bella bought stamps at the post office. Some of the stamps had a snowflake design, some had a truck design, and some had a rose design. Bella bought 11 snowflake stamps. She bought 9 more truck stamps than snowflake stamps, and 13 fewer rose stamps than truck stamps. How many stamps did Bella buy in all? & The number of truck stamps is 11 + 9 = 20. The number of rose stamps is 20 − 13 = 7. Bella bought 11 + 20 + 7 = 38 stamps in all. \\
            \midrule
            TimeQA & Who was the spouse of Anna Karina from 1968 to 1974? & Anna Karina's spouse from 1968 to 1974 was French actor Pierre Fabre. \\
            \midrule
            SituatedQA & what is the biggest hotel in las vegas nv as of November 28, 1993 & The biggest hotel in Las Vegas, Nevada as of November 28, 1993 was the Excalibur Hotel \& Casino. \\
            \midrule
            MuSiQue & What year saw the creation of the region where the county of Hertfordshire is located? & 1994 \\
            \midrule
            StrategyQA & Would a Monoamine Oxidase candy bar cheer up a depressed friend? & A Monoamine Oxidase (MAO) candy bar would not cheer up a depressed friend. MAO is an enzyme that breaks down neurotransmitters such as serotonin, dopamine, and norepinephrine. These neurotransmitters are associated with mood, so inhibiting MAO can lead to an increase in their levels. \\
            \midrule
        \bottomrule
        \end{tabular}
        \caption{Demonstration exemplar for standard prompting.
        }
        \label{tab:baseline_exemplars}
\end{table*}

\begin{table*}
    \centering
    \begin{tabular}{p{1.5cm}p{4.5cm}p{7cm}}
        \toprule
         dataset & Question & Answer\\
      \midrule
            MMLU Physics & A point charge, Q1 = +1 mC, is fixed at the origin. How much work is required to move another charge, Q2 = +8 $\mu$C, from the point (0, 4 meters) to the point (3 meters, 0)? \newline 
            0) 3.5 J \newline
            1) 6.0 J \newline
            2) 22.5 J \newline
            3) 40 J & Sure, let's think step by step. \newline
            
            1. Calculate the electric potential energy of Q2 at the initial position. The electric potential energy (U) of a point charge Q2 at a distance r from another point charge Q1 is given by the following equation: U = k * Q1 * Q2 / r, where k is Coulomb's constant.\newline
            
            Therefore, the electric potential energy of Q2 at the initial position is:\newline
            $U_\text{initial}$ = $(9 \times 10^9 N m^2 / C^2) * (1 \times 10^{-3} C) * (8 \times 10^{-6} C) / 4 m$ = 18 J\newline
            
            2. Calculate the electric potential energy of Q2 at the final position. The electric potential energy of Q2 at the final position is:\newline
            $U_\text{final}$ = $(9 \times 10^9 N m^2 / C^2) * (1 \times 10^{-3} C) * (8 \times 10^{-6} C) / 3 m$ = 24 J\newline
            
            3. Calculate the work required to move Q2 from the initial position to the final position. The work (W) required to move a charge from one position to another is equal to the change in electric potential energy ($\Delta$U) between the two positions.\newline
            W = $\Delta$U = $U_\text{final}$ - $U_\text{initial}$ = 24 J - 18 J = 6 J\newline
            
            Therefore, the answer is 1) 6.0 J. \\
            \midrule
        \bottomrule
        \end{tabular}
        \caption{Demonstration exemplar for CoT prompting for MMLU Physics.
        }
        \label{tab:cot_exemplars_physics}
\end{table*}

\begin{table*}
    \centering
    \begin{tabular}{p{1.5cm}p{4.5cm}p{7cm}}
        \toprule
         dataset & Question & Answer\\
      \midrule
            MMLU Chemistry & A sample of an unknown chloride compound was dissolved in water, and then titrated with excess Pb(NO3)2 to create a precipitate. After drying, it is determined there are 0.0050 mol of precipitate present. What mass of chloride is present in the original sample?\newline
            0) 0.177 g\newline
            1) 0.355 g\newline
            2) 0.522 g\newline
            3) 0.710 g & Here's the step-by-step solution:\newline
            Identify the precipitate: The precipitate formed when a chloride compound reacts with lead nitrate (Pb(NO3)2) is lead chloride (PbCl2).\newline
            
            Determine the mole ratio: The balanced chemical equation for the reaction between a chloride compound (represented as MCl, where M is the unknown metal) and lead nitrate is: $2MCl$ + $Pb(NO_3)_2 \longrightarrow 2MNO_3$ + $PbCl_2$
            \newline
            From the balanced equation, you can see that 2 moles of chloride ions (Cl-) react with 1 mole of lead nitrate to produce 1 mole of lead chloride precipitate. Additionally, 1 mole of lead chloride contains 2 moles of chloride ions.\newline
            
            Calculate the moles of chloride ions: Since 0.0050 moles of lead chloride precipitate were formed, there must have been 2 × 0.0050 = 0.010 moles of chloride ions in the original sample.\newline
            
            Calculate the mass of chloride ions: The molar mass of chlorine (Cl) is 35.45 g/mol. Therefore, the mass of 0.010 moles of chloride ions is:\newline
            mass = moles × molar mass = 0.010 mol × 35.45 g/mol = 0.355 g\newline
            
            Therefore, the mass of chloride present in the original sample is 0.355 g. The correct answer is option 1. \\
        \bottomrule
        \end{tabular}
        \caption{Demonstration exemplar for CoT prompting for MMLU Chemistry.
        }
        \label{tab:cot_exemplars_chem}
\end{table*}

\begin{table*}
    \centering
    \begin{tabular}{p{1.5cm}p{4.5cm}p{7cm}}
        \toprule
         dataset & Question & Answer\\
      \midrule
            GSM8K & Bella bought stamps at the post office. Some of the stamps had a snowflake design, some had a truck design, and some had a rose design. Bella bought 11 snowflake stamps. She bought 9 more truck stamps than snowflake stamps, and 13 fewer rose stamps than truck stamps. How many stamps did Bella buy in all? & Here are the steps to solve the problem:\newline
            
            Step 1: Find the number of truck stamps. Adding 9 to the number of snowflake stamps gives the number of truck stamps: 11 + 9 = 20\newline
            
            Step 2: Find the number of rose stamps. Subtracting 13 from the number of truck stamps gives the number of rose stamps: 20 - 13 = 7\newline
            
            Step 3: Add the number of snowflake, truck, and rose stamps to find the total number of stamps Bella bought: 11 + 20 + 7 = 38\newline
            
            Solution: Bella bought 38 stamps in all. \\
        \bottomrule
        \end{tabular}
        \caption{Demonstration exemplar for CoT prompting for GSM8K.
        }
        \label{tab:cot_exemplars_gsm8k}
\end{table*}

\subsection{Chain of Thought (CoT)}
\label{app:cot_prompt}
For zero-shot CoT prompting, we simply append \textit{Let's think step by step.} to the question to query the model.

For few-shot CoT prompting, we use the same template as the Baseline prompting in Sec.~\ref{app:baseline_prompt} by replacing the few-shot examples using CoT responses, as shown in Tables~\ref{tab:cot_exemplars_physics}, \ref{tab:cot_exemplars_chem}, \ref{tab:cot_exemplars_gsm8k}, \ref{tab:cot_exemplars_knowledge}, and \ref{tab:cot_exemplars_multihop}.

\begin{table*}
    \centering
    \begin{tabular}{p{1.5cm}p{4.5cm}p{7cm}}
        \toprule
         dataset & Question & Answer\\
      \midrule
           TimeQA & Who was the spouse of Anna Karina from 1968 to 1974? & Step 1: Identify Anna Karina's spouses:Jean-Luc Godard (1961–1965).\newline
           Pierre Fabre (1968–1974)\newline
           Daniel Duval (1978–1981)\newline
           Dennis Berry (1982–1994)\newline
           
           Step 2: Determine which spouse was married to Anna Karina from 1968 to 1974.\newline
           Jean-Luc Godard was married to Anna Karina from 1961 to 1965, so he is not the correct answer.\newline
           Pierre Fabre was married to Anna Karina from 1968 to 1974, so he is the correct answer.\newline
           Daniel Duval was married to Anna Karina from 1978 to 1981, so he is not the correct answer.\newline
           Dennis Berry was married to Anna Karina from 1982 to 1994, so he is not the correct answer.\newline
           
           Therefore, Anna Karina's spouse from 1968 to 1974 was Pierre Fabre. \\
            \midrule
            SituatedQA & what is the biggest hotel in las vegas nv as of November 28, 1993 & Sure, let's think step by step:

What were the largest hotels in Las Vegas in 1993?
According to a 1993 article in the Las Vegas Review-Journal, the largest hotels in Las Vegas at the time were:

MGM Grand (5,005 rooms)
Excalibur (4,008 rooms)
The Mirage (3,049 rooms)
Flamingo Hilton (3,565 rooms)
Caesars Palace (2,577 rooms)
Which of these hotels was the largest on November 28, 1993?
The MGM Grand opened in December 1993, so it would not have been the largest hotel on November 28, 1993. The Excalibur opened in June 1990, so it would have been the largest hotel on November 28, 1993.

Therefore, the largest hotel in Las Vegas on November 28, 1993 was the Excalibur. \\
            \midrule
        \bottomrule
        \end{tabular}
        \caption{Demonstration exemplar for CoT prompting for TimeQA and SituatedQA.
        }
        \label{tab:cot_exemplars_knowledge}
\end{table*}

\begin{table*}
    \centering
    \begin{tabular}{p{1.5cm}p{4.5cm}p{7cm}}
        \toprule
         dataset & Question & Answer\\
      \midrule
            MuSiQue & What year saw the creation of the region where the county of Hertfordshire is located? & The county of Hertfordshire is in the 'East of England' region. The 'East of England' region was created in 1994. So, the answer is 1994. \\
            \midrule
            StrategyQA & Would a Monoamine Oxidase candy bar cheer up a depressed friend? & No, a Monoamine Oxidase (MAO) candy bar would not cheer up a depressed friend. In fact, it could be dangerous for them to consume. Here's a step-by-step explanation:

1. MAO is an enzyme that breaks down neurotransmitters like serotonin, dopamine, and norepinephrine. These neurotransmitters are important for regulating mood, and low levels of them are associated with depression.

2. MAO inhibitors (MAOIs) are a type of medication that blocks the action of MAO, which can increase levels of these neurotransmitters in the brain. MAOIs are sometimes used to treat depression, but they can have serious side effects, including interactions with certain foods and medications.

3. A MAO candy bar would not contain enough MAOI to have a therapeutic effect on depression. In fact, it is unlikely that a candy bar could contain any MAOI at all, as these medications are typically taken in pill form.

In conclusion, a MAO candy bar would not cheer up a depressed friend and could even be dangerous for them to consume.  \\
            \midrule
        \bottomrule
        \end{tabular}
        \caption{Demonstration exemplar for CoT prompting for MuSiQue and StrategyQA.
        }
        \label{tab:cot_exemplars_multihop}
\end{table*}

\subsection{Take a Deep Breath (TDB)}
\label{app:tdb_prompt}
We study the zero-shot prompting found in~\cite{yang2023large}: we take \textit{Take a deep breath and work on this problem step-by-step}, and prepend it to the question.

\section{Examples of Error Analysis and Wins of \name{}}
\label{app:errors_wins}
\subsection{MMLU error analysis}\label{app:mmlu_errors}
In Tables~\ref{tab:mmlu_factual_error}-\ref{tab:mmlu_reasoning_error}, we show one example for each of the 5 error categories we identified through error analysis on \name{}.
\begin{table*}
    \centering
    \begin{tabular}{p{2cm}p{11cm}}
        \toprule
         Question & An alarm whose frequency is 400 Hz is dropped out of a third-floor window. The student who drops it measures the frequency with a very sensitive oscilloscope. The measured frequency\newline
         0) appears higher than 400 Hz and the frequency increase as it falls\newline
         1) appears higher than 400 Hz and the frequency decreases as it falls\newline
         \textcolor{green}{2) appears lower than 400 Hz and the frequency decreases as it falls}\newline
         3) appears lower than 400 Hz and the frequency increases as it falls\\
      \toprule
      Principles & Doppler Effect: the effect produced by the motion of a wave source, observer, or both that causes a change in wave frequency perceived by an observer relative to the frequency produced by the source. The Doppler effect is observed with all types of waves, including electromagnetic and acoustic waves. It was named after Austrian physicist Christian Doppler. \\
      \toprule
      Answer & Using the Principles of Doppler Effect, we can solve the problem as following: \newline
      
      Step 1: Apply Doppler Effect to find out the measured frequency.\newline
      The Doppler Effect is the change in frequency of a wave as measured by an observer who is moving relative to the source of the wave. In this case, the source of the wave is the alarm, and the observer is the student who drops it.\newline
      
      \textcolor{red}{As the alarm falls, the distance between the alarm and the student decreases.} This means that the frequency of the wave as measured by the student will increase. Therefore, the measured frequency will appear higher than 400 Hz, and the frequency will increase as the alarm falls.\newline
      
      Therefore, the correct answer is 0) appears higher than 400 Hz and the frequency increase as it falls. \\
        \bottomrule
        \end{tabular}
        \caption{Error example of \name{} on MMLU high-school Physics in the class of \textbf{Factual Error}. In this case, the Factual error leads to the wrong answer despite that the Principle and Reasoning are all sound.
        }
        \label{tab:mmlu_factual_error}
\end{table*}

\begin{table*}
    \centering
    \begin{tabular}{p{2cm}p{11cm}}
        \toprule
         Question & An astronaut lands on a planet whose mass and radius are each twice that of Earth. If the astronaut weighs 800 N on Earth, how much will he weigh on this planet?\newline
         0) 200 N\newline
         \textcolor{green}{1) 400 N}\newline
         2) 800 N\newline
         3) 1600 N\\
      \toprule
      Principles & Weight: W = m * g, where W is the weight, m is the mass and g is the acceleration due to gravity\newline
      Acceleration due to gravity: g = G * m / $r^2$, where G is the gravitational constant, m is the mass of the object and r is the radius of the object \\
      \toprule
      Answer & Using the Principles of Newton's law of universal gravitation and Weight, we can solve the problem as following: \newline
      
      Step 1: Apply Newton's law of universal gravitation to find out the acceleration due to gravity on the planet.\newline
      The acceleration due to gravity on a planet is given by the formula:\newline
      g = G * m / $r^2$, where g is the acceleration due to gravity, G is the gravitational constant, m is the mass of the planet, and r is the radius of the planet.\newline
      
      In this problem, we are given the following information:\newline
      m = 2 * $m_{earth}$\newline
      r = 2 * $r_{earth}$\newline
      
      Plugging these values into the formula, we get:\newline
      g = G * 2 * $m_{earth}$ / $(2 * r_{earth})^2$\newline
      \textcolor{red}{g = G * $m_{earth}$ / $r_{earth}^2$}\newline
      
      This is the same as the acceleration due to gravity on Earth.\newline
      
      Step 2: Apply Weight to find out the weight of the astronaut on the planet.\newline
      The weight of an object is given by the formula:\newline
      W = m * g, where W is the weight, m is the mass of the object, and g is the acceleration due to gravity.\newline
      
      In this problem, we are given the following information:\newline
      m = 800 N / $g_{earth}$\newline
      g = $g_{earth}$\newline
      
      Plugging these values into the formula, we get:\newline
      W = 800 N / $g_{earth} * g_{earth}$\newline
      W = 800 N\newline
      
      Therefore, the astronaut will weigh 800 N on the planet. The correct answer is 2) 800 N. \\
        \bottomrule
        \end{tabular}
        \caption{Error example of \name{} on MMLU high-school Physics in the class of \textbf{Math Error}. As highlighted, the math error leads to directly a wrong answer with a factor of 2 off.
        }
        \label{tab:mmlu_math_error}
\end{table*}

\begin{table*}
    \centering
    \begin{tabular}{p{2cm}p{11cm}}
        \toprule
         Question & Two spheres of net charge +5e and -6e briefly come into contact. Afterward, which of the following is a possible combination of net charges for the two spheres?\newline
         \textcolor{green}{0) +3e and -4e}\newline
         1) +1.5e and -2.5e\newline
         2) Both of the above\newline
         3) None of the above\\
      \toprule
      Principles & Law of conservation of charge: the total charge in a closed system remains constant.\newline
      \textcolor{red}{Coulomb's Law: the mutual repulsion of like charges demands that the charges be as far apart as possible.} \\
      \toprule
      Answer & Using the Principles of Conservation of charge, we can solve the problem as following: \newline
      
      Step 1: Apply Conservation of charge to find out the possible combination of net charges for the two spheres.\newline
      The total charge of the two spheres before they come into contact is +5e - 6e = -1e. After they come into contact, the total charge of the two spheres must still be -1e.\newline
      
      Step 2: Find out the possible combination of net charges for the two spheres.\newline
      The possible combination of net charges for the two spheres are:\newline
      (1) +3e and -4e\newline
      \textcolor{red}{(2) +1.5e and -2.5e}\newline
      
      Therefore, the correct answer is 2) Both of the above. \\
        \bottomrule
        \end{tabular}
        \caption{Error example of \name{} on MMLU high-school Physics in the class of \textbf{Principle Error}. In this case, Coulomb's Law is not needed for solving this problem. Instead, the Charge Quantization Principle is missing from the retrieved principles, and is in fact needed to rule out option (2) of fractional charges.
        }
        \label{tab:mmlu_principle_error}
\end{table*}

\begin{table*}
    \centering
    \begin{tabular}{p{2cm}p{11cm}}
        \toprule
         Question & The circumference of a helium-filled balloon is measured for three different conditions: at room temperature, after being in a warm oven for 30 minutes, and after being in a freezer for 30 minutes. A student plotting the circumference cubed $C^3$ as a function of temperature T, should expect to find which of the following?\newline
         0) a cubic relationship between $C^3$ and T\newline
         1) an indirect relationship between $C^3$ and T\newline
         2) an extrapolated temperature T where $C^3$ reaches zero\newline
         3) a maximum $C^3$ as the temperature T increases\\
      \toprule
      Principles & Ideal Gas Law: PV = nRT, where P is the pressure, V is the volume, n is the number of moles of gas, R is the ideal gas constant, T is the absolute temperature.\newline
      Volume of a sphere: $V = 4 / 3 * \pi * r^3$, where r is the radius.\newline
      Absolute temperature: Absolute temperature is always written as kelvin.\newline
      Absolute zero: Absolute zero is the lowest limit of the kelvin scale and is the lowest possible temperature (- 273.15 \textdegree C). \\
      \toprule
      Answer & Using the Principles of Charles's Law and Gay-Lussac's Law, we can solve the problem as following: \newline
      Step 1: Apply Charles's Law to find out the relationship between the volume and temperature of the gas. Charles's Law states that the volume of a fixed mass of a gas is directly proportional to the absolute temperature of the gas when pressure is kept constant. This means that if the temperature of the gas increases, the volume of the gas will also increase.\newline
      
      Step 2: Apply Gay-Lussac's Law to find out the relationship between the pressure and temperature of the gas. Gay-Lussac's Law states that the pressure of a fixed mass of a gas is directly proportional to the absolute temperature of the gas when volume is kept constant. This means that if the temperature of the gas increases, the pressure of the gas will also increase.\newline
      
      Step 3: Apply Charles's Law and Gay-Lussac's Law to find out the relationship between the circumference and temperature of the gas.The circumference of the balloon is directly proportional to the volume of the gas. This is because the circumference of the balloon is the distance around the outside of the balloon, and the volume of the gas is the amount of space that the gas takes up.\newline
      
      The pressure of the gas is directly proportional to the temperature of the gas. This is because the pressure of the gas is the force that the gas exerts on the walls of the balloon, and the temperature of the gas is the measure of the average kinetic energy of the gas molecules.\newline
      
      Therefore, the circumference of the balloon is directly proportional to the temperature of the gas. This means that if the temperature of the gas increases, the circumference of the balloon will also increase.\newline
      
      Step 4: Apply the relationship between the circumference and temperature of the gas to find out the relationship between the circumference cubed and temperature of the gas. The circumference cubed is directly proportional to the temperature of the gas. This is because the circumference cubed is the volume of the balloon, and the volume of the balloon is directly proportional to the temperature of the gas.\newline
      
      \textcolor{red}{Therefore, the student should expect to find a cubic relationship between $C^3$ and T.} The correct answer is 0) a cubic relationship between $C^3$ and T. \\
        \bottomrule
        \end{tabular}
        \caption{Error example of \name{} on MMLU high-school Physics in the class of \textbf{Context Loss}. In this case, at the end of the Reasoning chain, the model forgot the original question, and lose the context to apply the reasoning to the question.
        }
        \label{tab:mmlu_context_loss}
\end{table*}

\begin{table*}
    \centering
    \begin{tabular}{p{2cm}p{11cm}}
        \toprule
         Question & Physics students are checking the constant acceleration equations of kinematics by measuring the velocity of a tennis ball that is dropped and falls 6 meters and then passes through a photogate. The predicted velocity is $20\%$ above the velocity measured by the photogate. Which of the following best describes the cause of the large percent difference?\newline
         0) The ball changes its shape while falling.\newline
         1) The acceleration of gravity varies as the ball is falling.\newline
         2) Air resistance increases the acceleration of the ball.\newline
         \textcolor{green}{3) The acceleration of the balls varies with the velocity.}\\
      \toprule
      Principles & Vertical motion: y = 0.5 * g * $t^2$, where y is the vertical displacement, g is the acceleration due to gravity, t is the time. \newline
      Free fall: is a special case of motion with constant acceleration, in which the acceleration is due to gravity.\newline
      Acceleration to Velocity Formula: v = u + a * t, where v is the final speed, u is the initial velocity, a is the acceleration, and t is the time. \\
      \toprule
      Answer & Using the Principles of Vertical motion and Horizontal motion, we can solve the problem as following: \newline
      
      Step 1: Apply the Vertical motion equation to find the time it takes for the ball to fall 6 meters.\newline
      y = 0.5 * g * $t^2$\newline
      6 m = 0.5 * 9.8 m/$s^2$ * $t^2$\newline
      t = 1.12 s\newline
      
      Step 2: Apply the Horizontal motion equation to find the velocity of the ball when it passes through the photogate.\newline
      x = v * t\newline
      v = x / t\newline
      v = 6 m / 1.12 s\newline
      v = 5.36 m/s\newline
      
      Step 3: Compare the predicted velocity with the velocity measured by the photogate.
      The predicted velocity is $20\%$ above the velocity measured by the photogate, so the predicted velocity is 1.2 * 5.36 m/s = 6.43 m/s.
      
      Step 4: Find the cause of the large percent difference.
      The predicted velocity is 6.43 m/s, but the velocity measured by the photogate is 5.36 m/s. The difference is 1.07 m/s. This difference is caused by air resistance. Air resistance is a force that opposes the motion of an object through the air. It is caused by the friction between the object and the air molecules. The greater the surface area of the object, the greater the air resistance. The tennis ball has a large surface area, so it experiences a lot of air resistance. \textcolor{green}{This air resistance slows down the tennis ball and causes the difference between the predicted velocity and the velocity measured by the photogate.}
      
      \textcolor{red}{Therefore, the correct answer is 0) The ball changes its shape while falling.} \\
        \bottomrule
        \end{tabular}
        \caption{Error example of \name{} on MMLU high-school Physics in the class of \textbf{Reasoning Error}. In this case, the model was able to correctly identify air resistance is the cause of the velocity difference, but made the error at the final reasoning step of deriving the right answer.
        }
        \label{tab:mmlu_reasoning_error}
\end{table*}

\subsection{Example Wins from Step-Back Prompting}\label{app:stepback_wins}

Tables ~\ref{tab:mmluphysics_examples}-~\ref{tab:mmluphysics_examples_3}, ~\ref{tab:mmluchem_examples} ~\ref{tab:timeqa_examples}, ~\ref{tab:situatedqa_examples} and ~\ref{tab:strategyqa_examples} illustrate the some successful examples of \name{} on MMLU-Physics, MMLU-Chemistry, TimeQA, SituatedQA, and StrategyQA respectively.

\begin{table*}[t]
    \centering
    \begin{tabular}{p{3.3cm}p{10cm}}
        \toprule
         dataset & example \\
      \midrule
            MMLU-Physics & \textbf{Original Question}: What happens to the pressure, P, of an ideal gas if the temperature is increased by a factor of 2 and the volume is increased by a factor of 8 ? \newline
            0) P decreases by a factor of 16. \newline
            1) P decreases by a factor of 4.  \newline
            2) P decreases by a factor of 2.  \newline
            3) P increases by a factor of 4. \newline
            \textbf{Original Answer [Incorrect]}: The correct answer is 0. The ideal gas law is PV = nRT. If the temperature is increased by a factor of 2 and the volume is increased by a factor of 8, \textcolor{red}{then the pressure will decrease by a factor of 16}. \newline \textbf{Stepback Question}: What are the physics principles behind this question? \newline
            \textbf{Principles}: \newline
            Ideal gas law: PV = nRT, where P is the pressure, V is the volume, n is the number of moles, R is the gas constant, and T is the temperature.\newline
            \textbf{Final Answer}: Using the Principles of Ideal gas law, we can solve the problem as following: \newline
            Step 1: Apply Ideal gas law to find out the pressure of an ideal gas.
            The ideal gas law states that PV = nRT, where P is the pressure, V is the volume, n is the number of moles, R is the gas constant, and T is the temperature. \newline

            Step 2: Apply Ideal gas law to find out the pressure of an ideal gas when the temperature is increased by a factor of 2 and the volume is increased by a factor of 8.
            If the temperature is increased by a factor of 2, then T becomes 2T. If the volume is increased by a factor of 8, then V becomes 8V. \newline

            Substituting these values into the ideal gas law, we get:
            P(8V) = nR(2T) \newline

            Dividing both sides by 8V, we get:
            P = nR(2T) / 8V  \newline

            We can see that the pressure has decreased by a factor of 4. \newline

            Therefore, the correct answer is 1) P decreases by a factor of 4.\\
        \bottomrule
        \end{tabular}
        \caption{Illustration of wins of \name{} on the MMLU-Physics dataset.
        }
        \label{tab:mmluphysics_examples}
\end{table*}

\begin{table*}[t]
    \centering
    \begin{tabular}{p{3.3cm}p{10cm}}
        \toprule
         dataset & example \\
            \midrule
            MMLU-Physics & \textbf{Original Question}: A liquid flows at a constant flow rate through a pipe with circular cross-sections of varying diameters. At one point in the pipe, the diameter is 2 cm and the flow speed is 18 m/s. What is the flow speed at another point in this pipe, where the diameter is 3 cm?  \newline
            0) 4 m/s  \newline
            1) 6 m/s  \newline
            2) 8 m/s  \newline
            3) 12 m/s \newline 
            \textbf{Original Answer [Incorrect]}: The answer is 3) 12 m/s A step-by-step solution is below: The flow rate is constant, so the product of the speed and the cross-sectional area is constant. The cross-sectional area is proportional to the square of the diameter, so the speed is inversely proportional to the square of the diameter. The ratio of the speeds is the inverse of the ratio of the squares of the diameters, so the speed at the second point is \textcolor{red}{$18 m/s * (2 cm / 3 cm)^2$ = 12 m/s}. \newline 
            \textbf{Stepback Question}: What are the physics principles behind this question? \newline
            \textbf{Principles}: \newline
            Conservation of Mass: the amount of mass flowing through the pipe at two different points must be equal, or A1 * v1 = A2 * v2, where A1 is the area of cross-section one, v1 is the velocity at point 1, A2 is the area of cross-section two, v2 is the velocity at point 2. \newline
            Area of Circle: $A = \pi * r^2$, where r is the radius of the circle. \newline
            \textbf{Final Answer}: Using the Principles of Continuity equation, we can solve the problem as following:  \newline
            Step 1: Apply the Continuity equation to find out the flow speed at another point in this pipe. The continuity equation states that the product of the area and the velocity is constant at all points in a pipe. In this problem, we are given the following information: \newline
            $A1 = \pi * (2 cm)^2 = 4  \pi cm^2$ \newline
            v1 = 18 m/s \newline
            $A2 = \pi * (3 cm)^2 = 9 \pi cm^2$ \newline  \newline
            We can use the continuity equation to find v2: \newline
            A1 * v1 = A2 * v2 \newline
            $4\pi cm^2 * 18 m/s = 9\pi cm^2 * v2$ \newline
            v2 = 8 m/s \newline  \newline
            Therefore, the flow speed at another point in this pipe is 8 m/s. The correct answer is 2) 8 m/s. \\
        \bottomrule
        \end{tabular}
        \caption{Illustration of wins of \name{} on the MMLU-Physics dataset.
        }
        \label{tab:mmluphysics_examples_2}
\end{table*}

\begin{table*}[t]
    \centering
    \begin{tabular}{p{2.3cm}p{11cm}}
        \toprule
         dataset & example \\
            \midrule
            MMLU-Physics & \textbf{Original Question}: A 2 $\mu$F capacitor is connected directly to a battery. When the capacitor is fully charged, it stores 600 $\mu$C of charge. An experimenter replaces the 2 $\mu$F capacitor with three 18 $\mu$F capacitors in series connected to the same battery. Once the capacitors are fully charged, what charge is stored on each capacitor?  \newline 
            0) 100 $\mu$C  \newline 
            1) 200 $\mu$C  \newline 
            2) 600 $\mu$C  \newline 
            3) 1800 $\mu$C \newline 
            \textbf{Original Answer [Incorrect]}: \textcolor{red}{The answer is 100 $\mu$C. The total capacitance of the three 18 $\mu$F capacitors in series is 1 / (1 / 18 + 1 / 18 + 1 / 18) = 6 $\mu$F. The charge stored on the three capacitors in series is 6 $\mu$F * V = 600 $\mu$C. The charge stored on each capacitor is 600 $\mu$C / 3 = 200 $\mu$C}. \newline 
            \textbf{Stepback Question}: What are the physics principles behind this question? \newline
            \textbf{Principles}: \newline
            Capacitors: Q = C *V, where Q is the charge, C is the capacitance and V is the voltage. \newline Capacitors in series: C = 1 / (1/C1 + 1/C2 + ... + 1/Cn), where C is the equivalent capacitance and C1, C2, Cn are the capacitances of each capacitor. \newline
            Capacitor in parallel: C = C1 + C2 + ... + Cn, where C is the equivalent capacitance and C1, C2, Cn are the capacitances of each capacitor. \newline
            \textbf{Final Answer}: Using the Principles of Capacitance and Capacitors in series, we can solve the problem as following: \newline 
            Step 1: Apply Capacitance to find out the voltage of the battery.\newline 
            We know that the capacitance of the 2 $\mu$F capacitor is 2 $\mu$F and the charge stored on the capacitor is 600 $\mu$C. Using the formula for capacitance, we can find the voltage of the battery: \newline 
            C = Q / V \newline 
            2 $\mu$F = 600 $\mu$C / V \newline 
            V = 300 V \newline
            
            Step 2: Apply Capacitors in series to find out the equivalent capacitance of the three 18 $\mu$F capacitors in series. \newline 
            We know that the capacitance of each of the three capacitors is 18 $\mu$F. Using the formula for capacitors in series, we can find the equivalent capacitance of the three capacitors: \newline 
            1 / $C_{eq}$ = 1 / $C_1$ + 1 / $C_2$ + 1 / $C_3$  \newline 
            1 / $C_{eq}$ = 1 / 18 $\mu$F + 1 / 18 $\mu$F + 1 / 18 $\mu$F \newline 
            1 / $C_{eq}$ = 3 / 18 $\mu$F \newline 
            $C_{eq}$ = 6 $\mu$F \newline  \newline 
            Step 3: Apply Capacitance to find out the charge stored on each capacitor. \newline 
            We know that the equivalent capacitance of the three capacitors is 6 $\mu$F and the voltage of the battery is 300 V. Using the formula for capacitance, we can find the charge stored on each capacitor: \newline 
            C = Q / V  \newline 
            6 $\mu$F = Q / 300 V \newline 
            Q = 1800 $\mu$C \newline  \newline 
            Therefore, the charge stored on each capacitor is 1800 $\mu$C. The correct answer is 3) 1800 $\mu$C. \\
        \bottomrule
        \end{tabular}
        \caption{Illustration of wins of \name{} on the MMLU-Physics dataset.
        }
        \label{tab:mmluphysics_examples_3}
\end{table*}

\begin{table*}[t]
    \centering
    \begin{tabular}{p{2.3cm}p{11cm}}
        \toprule
         dataset & example \\
            \midrule
            MMLU-Chemistry & \textbf{Original Question}: Which of the following indicates that a reaction is spontaneous?\newline
            0) at equilibrium there are more products than reactants\newline
            1) the value of $\Delta G^0$ is greater than zero\newline
            2) the value of $\Delta S^0$ is greater than zero\newline
            3) the value of Keq is less than one\newline
            \textbf{Original Answer [Incorrect]}: \textcolor{red}{2) the value of $\Delta$S° is greater than zero.\newline
            A reaction is spontaneous if the value of $\Delta$S° is greater than zero. This is because a positive $\Delta$S° indicates that the reaction results in an increase in entropy, which is a measure of disorder. A spontaneous reaction is one that occurs without the input of energy, and an increase in entropy is a natural tendency of the universe.} \newline 
            \textbf{Stepback Question}: What are the chemistry principles behind this question? \newline
            \textbf{Principles}: \newline
            Spontaneous reactions: A spontaneous reaction is a reaction that occurs without the input of energy. \newline
            \textbf{Final Answer}: A reaction is spontaneous if the change in Gibbs free energy is negative ($\Delta G < 0$). The value of $\Delta G^0$ is greater than zero indicates that the reaction is not spontaneous.\newline
            
            The correct answer is 0) at equilibrium there are more products than reactants. \\
            
            \midrule
            MMLU-Chemistry & \textbf{Original Question}: Potassium-40 is a minor isotope found in naturally occurring potassium. It is radioactive and can be detected on simple radiation counters. How many protons, neutrons, and electrons does potassium-40 have when it is part of K2SO4?\newline
            0) 21 neutrons, 19 protons, 18 electrons\newline
            1) 20 neutrons, 19 protons, 19 electrons\newline
            2) 21 neutrons, 19 protons, 19 electrons\newline
            3) 19 neutrons, 19 protons, 19 electrons\newline
            \textbf{Original Answer [Incorrect]}: \textcolor{red}{The correct answer is 2).\newline 
            Potassium-40 has 19 protons and 21 neutrons. When it is part of K2SO4, it has 19 electrons.} \newline 
            \textbf{Stepback Question}: What are the chemistry principles behind this question? \newline
            \textbf{Principles}: \newline
            Atomic number: The atomic number of an element is the number of protons in the nucleus of an atom of that element. \newline
            \textbf{Final Answer}: The atomic number of potassium is 19, so all potassium atoms have 19 protons. The mass number of potassium-40 is 40, so it has 40 - 19 = 21 neutrons. When potassium is part of a compound, it loses one electron to become a potassium ion. Therefore, potassium-40 has 19 - 1 = 18 electrons when it is part of K2SO4.\newline
            The correct answer is 0) 21 neutrons, 19 protons, 18 electrons.\\
        \bottomrule
        \end{tabular}
        \caption{Illustration of wins of \name{} on the MMLU-Chemistry dataset.
        }
        \label{tab:mmluchem_examples}
\end{table*}

\begin{table*}
    \centering
    \begin{tabular}{p{3.3cm}p{10cm}}
        \toprule
         dataset & example \\
      \midrule
            TimeQA & \textbf{Original Question}: Carlos Ascues played for which team from 2013 to 2014? \newline \textbf{Original Answer [Incorrect]}: Carlos Ascues played for \textcolor{red}{Benfica} from 2013 to 2014. \newline \textbf{Stepback Question}: Which teams did Carlos Ascues play for? \newline
            \textbf{Final Answer [Correct]}: From 2013 to 2014, Carlos Ascues played for the Greek team Panetolikos.\\
            \midrule
            TimeQA & \textbf{Original Question}: Seth Nana Twumasi played for which team between Jan 2011 and Feb 2011? \newline \textbf{Original Answer [Incorrect]}: Seth Nana Twumasi played for the \textcolor{red}{FC Honka team} between Jan 2011 and Feb 2011. \newline \textbf{Stepback Question}: Which teams did Seth Nana Twumasi play for? \newline
            \textbf{Final Answer [Correct]}: Northampton. He was on a loan transfer from Peterborough.\\
            \midrule
            TimeQA & \textbf{Original Question}: What position did Carmel Hanna take in May 2002? \newline \textbf{Original Answer [Incorrect]}: According to the article, Carmel Hanna was appointed as \textcolor{red}{Minister of State for Children and the Family} in May 2002. She held this position until 2005. \newline \textbf{Stepback Question}: What positions did Carmel Hanna hold? \newline
            \textbf{Final Answer [Correct]}: From the context, we can see that Carmel Hanna was named minister of Employment and Learning (2001). We can then infer that she held the position of Minister for Employment and Learning in May 2002.\\
        \bottomrule
        \end{tabular}
        \caption{Illustration of wins of \name{} on the TimeQA dataset.
        }
        \label{tab:timeqa_examples}
\end{table*}

\begin{table*}
    \centering
    \begin{tabular}{p{3.3cm}p{10cm}}
        \toprule
         dataset & example \\
      \midrule
            SituatedQA & \textbf{Original Question}: when was the last time miss texas won miss america as of November 29, 2010 \newline \textbf{Original Answer [Incorrect]}: \textcolor{red}{Laura Leigh Covington was the last Miss Texas to win the Miss America competition. This happened in 2000.} \newline \textbf{Stepback Question}: which years did miss texas win miss america \newline
            \textbf{Final Answer [Correct]}: According to Wikipedia, the last time Miss Texas won Miss America was in 1975.\\
            \midrule
            SituatedQA & \textbf{Original Question}: when is the last time congress passed a complete budget as of 2021 \newline \textbf{Original Answer [Incorrect]}: \textcolor{red}{The last time Congress passed a complete budget as of 2021 was on January 19, 2018.} \newline \textbf{Stepback Question}: what are the years that congress passed a complete budget as of 2021 \newline
            \textbf{Final Answer [Correct]}: The last time Congress passed a complete budget as of 2021 was December 27, 2020, the Consolidated Appropriations Act, 2021.\\
            \midrule
            SituatedQA & \textbf{Original Question}: when was the last time unc did not make the ncaa tournament as of 2003 \newline \textbf{Original Answer [Incorrect]}: \textcolor{red}{The last time UNC did not make the NCAA tournament was 1990.}\newline \textbf{Stepback Question}: which years did unc not make the ncaa tournament as of 2003 \newline
            \textbf{Final Answer [Correct]}: The correct answer is 2002-03. The context shows the standings of North Carolina Tar Heels men's basketball. From the standings, we can see that this team has 19 wins and 16 losses. 19-16 is not good enough to make the NCAA tournament, which is mentioned in the context. From the standings, we can also see that this is the standing for 2002-03 season. So the last time UNC did not make the NCAA tournament as of 2003 is 2002-03.\\
        \bottomrule
        \end{tabular}
        \caption{Illustration of wins of \name{} on the SituatedQA dataset.
        }
        \label{tab:situatedqa_examples}
\end{table*}

\begin{table*}
    \centering
    \begin{tabular}{p{3.3cm}p{10cm}}
        \toprule
         dataset & example \\
      \midrule
            StrategyQA & \textbf{Original Question}: Could you drive a Rowe 550 to the 2008 Summer Olympics? \newline \textbf{Original Answer [Incorrect]}: The 2008 Summer Olympics was held in Beijing, China. The Rowe 550 is an all-electric car. The Rowe 550 has a range of 70 miles on a single charge and a top speed of about 30 mph. \textcolor{red}{The distance between Beijing and the U.K. is about 4000 miles.}\newline \textbf{Stepback Question}: What kind of vehicle is a Rowe 550? \newline
            \textbf{Final Answer [Correct]}: The Roewe 550 was released in 2008; the 2008 Summer Olympics were held from August 8-24, 2008. So you could drive a Rowe 550 to the 2008 Summer Olympics.\\
            \midrule
            StrategyQA & \textbf{Original Question}: what channel does thursday night football come on regular tv as of 2017? \newline \textbf{Original Answer [Incorrect]}: Thursday Night Football is a weekly series of National Football League (NFL) games broadcast on Thursday nights. It started in 2006 on NFL Network. \textcolor{red}{Under a new agreement between the NFL and CBS, CBS will air $50\%$ of the Thursday Night Football games and NFL Network will air the other $50\%$}. \newline \textbf{Stepback Question}: which channel is thursday night football on as of 2017? \newline
            \textbf{Final Answer [Correct]}: In 2017, Thursday Night Football was aired on CBS, NBC, and NFL Network.\\
            \midrule
            StrategyQA & \textbf{Original Question}: who is the present president of assam sahitya sabha as of 2018? \newline \textbf{Original Answer [Incorrect]}: \textcolor{red}{Dr. Nagen Saikia} \newline \textbf{Stepback Question}: who lead assam sahitya sabha as of 2018? \newline
            \textbf{Final Answer [Correct]}: The answer is Paramananda Rajbongshi. As per a news from 2019, The Asam Sahitya Sabha, which was established in 1917, in Sibsagar, Assam, had got Paramananda Rajbongshi as its president. So, as of 2018, the president of assam sahitya sabha is Paramananda Rajbongshi.\\
        \bottomrule
        \end{tabular}
        \caption{Illustration of wins of \name{} on the StrategyQA dataset.
        }
        \label{tab:strategyqa_examples}
\end{table*}

\end{document}